%% file: main.tex
\documentclass[11pt,a4paper]{article}

\usepackage[utf8]{inputenc}
\usepackage[T1]{fontenc}
\usepackage{lmodern}
\usepackage[a4paper,margin=25mm]{geometry}
\usepackage{amsmath,amssymb,mathtools}
\usepackage{graphicx}
\usepackage{booktabs,longtable,array,multirow,makecell,calc}
\usepackage{pdflscape}
\usepackage{caption}
\usepackage{float}
\usepackage[table]{xcolor}
\usepackage{soul}
\usepackage{textcomp}
\usepackage{microtype}
\usepackage[numbers,sort&compress]{natbib}
\usepackage{xurl}
\usepackage[hidelinks]{hyperref}

\graphicspath{{figures/}}
\renewcommand{\arraystretch}{1.12}
\sethlcolor{yellow!30}

\newcounter{supptext}

\title{Domain-Adaptive Pretraining Enhances Water Treatment Semantic Representation for Large-Scale Structured Literature Mining}
\author{\parbox{0.94\textwidth}{\centering
Mudi Zhai\textsuperscript{a},
Ruihong Qiu\textsuperscript{b},
Qingyun Zeng\textsuperscript{c,d},\\
T. David Waite\textsuperscript{a},
Bing-Jie Ni\textsuperscript{a},
Haoran Duan\textsuperscript{a,e,*}\\[0.8em]
\small \textsuperscript{a}UNSW Water Research Centre, School of Civil and Environmental Engineering, The University of New South Wales, Sydney, NSW 2052, Australia\\
\small \textsuperscript{b}School of Electrical Engineering and Computer Science, The University of Queensland, Brisbane, QLD 4072, Australia\\
\small \textsuperscript{c}Microsoft Copilot Studio AI, Redmond, WA 98052, United States\\
\small \textsuperscript{d}Departments of Mathematics \& Department of Computer and Information Science, University of Pennsylvania, Philadelphia, PA 19104, United States\\
\small \textsuperscript{e}Department of Civil Engineering, The University of Hong Kong, Pokfulam, Hong Kong SAR, China\\[0.5em]
\small \textsuperscript{*}Corresponding author: \href{mailto:haoran.duan@hku.hk}{haoran.duan@hku.hk}
}}
\date{}

\begin{document}
\maketitle

\begin{abstract}
Water treatment research is expanding rapidly, but much of the knowledge acquired from this research remains scattered across unstructured literature. The field still lacks a dedicated language model that can efficiently capture water treatment-specific domain semantics for large-scale literature mining. Here, we address this by developing WaterBERT (Bidirectional Encoder Representations from Transformers), a domain-adapted encoder model designed for semantic representation and structured information extraction from water treatment texts. WaterBERT was developed by continual pretraining on a large-scale water treatment corpus comprising about 2.97 billion tokens. Three fine-tuned models based on WaterBERT were systematically evaluated on downstream tasks, achieving the best overall performance among general-purpose and domain-specific BERT models, with F1 scores of 90.12\% for multiclass treatment process classification, 79.50\% for named entity recognition, and 74.04\% for relation extraction. Beyond these benchmark tasks, we further demonstrated WaterBERT's advantages for large-scale literature processing. Applied to 5,144 \emph{Environmental Science \& Technology} articles, WaterBERT--BERTopic identified coherent, diverse, and domain-specific research topics without predefined categories. Building on WaterBERT, we processed 693,211 abstracts at substantially lower cost than commercial LLMs while retaining competitive extraction performance to construct a structured water treatment knowledge graph. The knowledge graph was then integrated with lexical and dense retrieval to develop a Water Knowledge-Enhanced Retrieval System (WaterKERS), which achieved a relevance score of 77.7, substantially outperforming text-based retrieval baselines (54.7--64.5). Through WaterBERT, this study provides a compact and scalable semantic foundation for large-scale information processing and evidence mapping in water treatment research.
\end{abstract}


\input{main_body}
\clearpage
\input{supplementary}

\end{document}

%% file: main_body.tex
\section{Introduction}\label{sec:introduction}

Driven by increasingly strict discharge requirements, emerging contaminants, resource recovery objectives and broader sustainability goals, water and wastewater treatment has developed into a rapidly expanding interdisciplinary field, generating a large and diverse body of technical literature.\cite{ref01,ref02} By 2026, a Web of Science search for articles associated with wastewater and water treatment yielded more than two million records, including more than 700,000 in the Web of Science Core Collection. These publications contain extensive information on treatment processes, pollutant characteristics, material properties, operating conditions, and treatment performance. Systematically extracting and organizing this information could support the ongoing development and better management of water treatment systems. Potential applications include building training datasets for treatment performance models, creating structured evidence bases for evaluating treatment technologies, and developing domain knowledge graphs to support decision-making in water/wastewater treatment plants.\cite{ref03,ref04} However, much of this information remains dispersed across a vast and largely unstructured body of literature, making systematic extraction difficult.

Natural language processing (NLP) offers a pathway to transform unstructured scientific text into structured knowledge, but effective information extraction from water treatment literature requires domain-specific semantic interpretation rather than general text understanding alone.\cite{ref05,ref06,ref07} What a term means depends on its role in and/or relevance to the water treatment system. The same chemical may function as a target pollutant, dosed reagent, catalyst, electrolyte, nutrient source, or background matrix. For example, NaCl may be the target pollutant in studies on saline wastewater treatment or desalination, but it may also serve as an electrolyte or background matrix component in electrochemical treatment systems.\cite{ref08,ref09} Likewise, a reported removal efficiency becomes meaningful only when it is linked to the correct treatment process, target contaminant, and operating conditions, because removal may result from adsorption, degradation, membrane rejection, or biological transformation.\cite{ref10}

When used as standalone models with only simple prompting, general-purpose large language models (LLMs) are likely to perform inconsistently on specialised literature-extraction tasks in water treatment.\cite{ref06,ref11} More elaborate strategies, including prompt engineering, agent-based workflows, and Retrieval-Augmented Generation (RAG), can improve extraction performance in water-related applications.\cite{ref12,ref13,ref14} However, these methods often incur substantial costs. For example, Jiang et al.~\cite{ref13} reported that a structured-prompting pipeline cost approximately CNY 3,000 (\textasciitilde USD 417) to process 5,100 papers. Extrapolating this rate to a field-wide corpus of more than 700,000 water treatment documents would result in a total cost exceeding USD 57,000. Such costs may limit the feasibility of processing literature at this scale. Moreover, previous studies found that adapting open-source LLMs via domain-specific fine-tuning for water-treatment-related tasks offered only limited gains and even underperformed strong commercial models.\cite{ref06,ref15} Therefore, more efficient models that can be adapted to the water treatment domain and deployed for corpus-scale information extraction and organisation are needed.

Among NLP approaches, encoder models from the Bidirectional Encoder Representations from Transformers (BERT) family are well suited to this setting.\cite{ref16} Although LLMs are powerful for generation and synthesis, corpus-scale information extraction relies mainly on discriminative tasks such as document screening, named entity recognition, relation extraction, and semantic embedding for topic modelling. Continued Domain-Adaptive Pretraining (DAPT) enables BERT models to learn specialised terminology and usage while retaining general scientific knowledge.\cite{ref07} Previous studies have shown that domain-adapted BERT models can outperform general-purpose LLMs on tasks requiring specialised semantic understanding.\cite{ref17,ref18,ref19} These models also support efficient and reproducible processing of large document collections and have been applied to large-scale literature analysis in biomedicine,\cite{ref20} materials science,\cite{ref21} and climate change.\cite{ref22,ref23} For example, Zhang et al.~\cite{ref20} employed a PubMedBERT-based information extraction pipeline to process more than 34 million PubMed abstracts, constructing a large-scale knowledge graph comprising 10,686,927 unique entities and 30,758,640 unique relations. Shetty et al.~\cite{ref21} further showed that MaterialsBERT, trained on 2.4 million materials science abstracts, enabled the automatic extraction of approximately 300,000 material--property records from around 130,000 polymer abstracts. However, no comparable domain-adapted model has been developed for the specialised terminology and relational semantics of the water treatment literature.

Here, we develop and apply WaterBERT, the first domain-adapted encoder model for semantic representation and information extraction in water/wastewater treatment research. WaterBERT was developed through continual pretraining of SciBERT on a large-scale water treatment corpus comprising approximately 2.23 billion words. We evaluated WaterBERT across three downstream tasks, including text classification, named entity recognition (NER), and relation extraction (RE). We then applied WaterBERT to large-scale water treatment literature for two corpus-level analyses: topic modelling and structured knowledge graph extraction. The resulting knowledge graph was further integrated into a Water Knowledge-Enhanced Retrieval System (WaterKERS) and evaluated for its ability to enhance literature retrieval. By combining domain-adapted semantic representation with corpus-scale literature analysis, this work provides a practical platform for organizing, retrieving, and mapping evidence across the rapidly expanding water treatment literature.

\section{Methods}\label{sec:methods}

\subsection{Overall workflow}\label{sec:workflow}

\begin{figure}[t]
\centering
\includegraphics[width=0.98\linewidth]{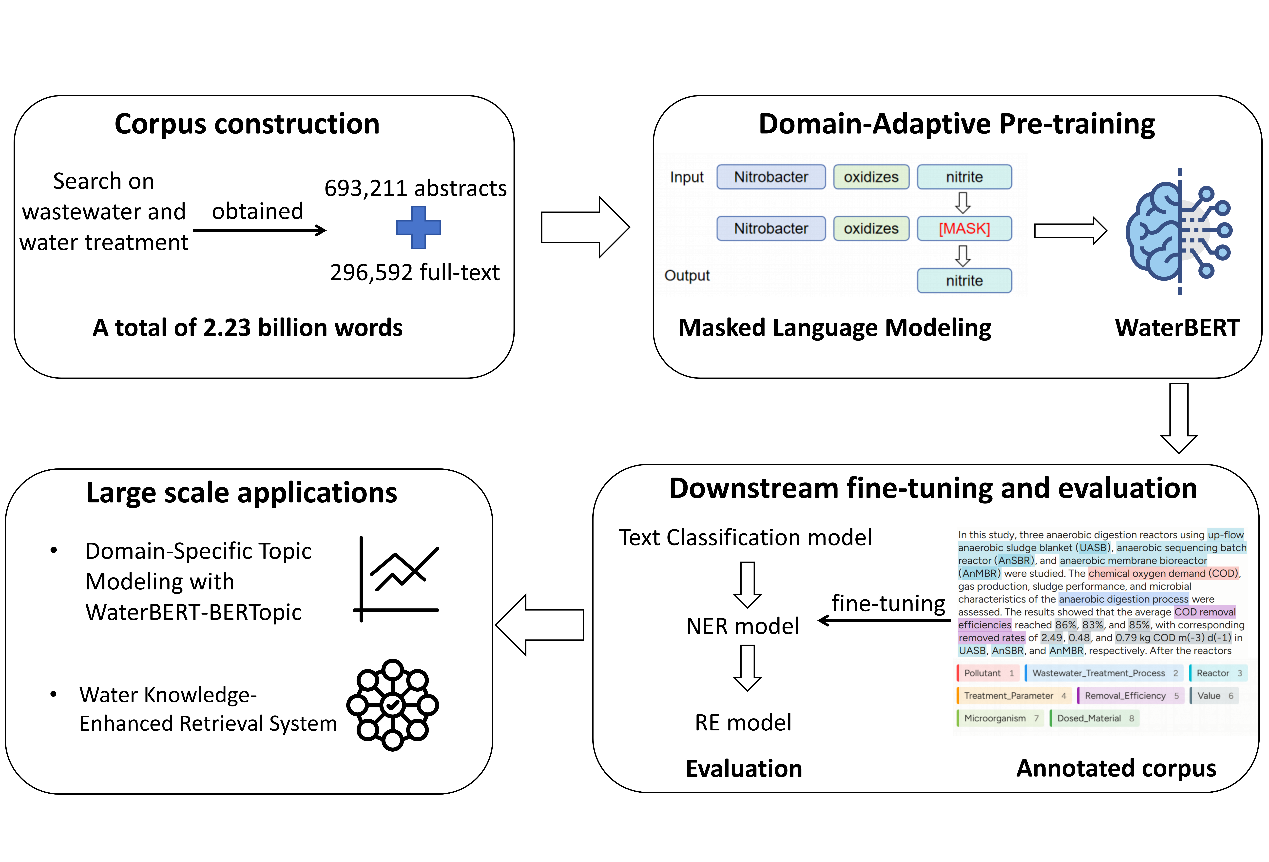}
\caption{Overall workflow of WaterBERT. A domain-specific corpus was constructed from water treatment literature and used for masked language model pretraining. WaterBERT was then fine-tuned for text classification, named entity recognition, and relation extraction, and applied to large-scale literature for topic analysis and structured knowledge graph construction.}
\label{fig:workflow}
\end{figure}

The overall workflow of this study is shown in Fig.~\ref{fig:workflow}. To develop a language model adapted to water treatment text, we first constructed a large-scale water treatment corpus from scientific literature. We then performed domain-adaptive pretraining (DAPT) on this corpus using the masked language modeling objective (MLM), in which masked tokens are predicted from their surrounding context, to obtain WaterBERT. After pretraining, WaterBERT was fine-tuned on annotated datasets for three downstream tasks: text classification, named entity recognition (NER), and relation extraction (RE). These tasks were used to evaluate whether WaterBERT can support both document-level literature screening and the extraction of wastewater treatment information. Finally, we examined the potential of WaterBERT for large-scale applications through two case studies: BERTopic-based topic modelling using WaterBERT embeddings and knowledge graph construction using the fine-tuned NER and RE models. Building on the resulting knowledge graph, we further developed a Water Knowledge-Enhanced Retrieval System (WaterKERS) and evaluated its ability to improve literature retrieval.

\subsection{Corpus construction}\label{sec:corpus}

A domain-specific corpus is essential for DAPT. We therefore constructed a dedicated corpus covering 693,211 water treatment publications as the pretraining foundation for WaterBERT. We searched the Web of Science Core Collection using the topics ``wastewater treatment'' and ``water treatment'', retrieving 693,211 article abstracts. The corpus included publications available up to February 2026. Among these publications, full texts were additionally available for 296,592 articles from Elsevier and were incorporated into the pretraining corpus. Table~\ref{tab:si1} presents the high-frequency terms in the resulting corpus. The five most frequent terms were ``water'', ``treatment'', ``surface'', ``removal'', and ``concentration'', indicating that the corpus was centered on water treatment related research. The raw texts were then deduplicated and filtered to retain only English content. The cleaned texts were segmented into chunks of up to 512 tokens, consistent with the maximum input length of BERT. The final corpus contained approximately 2.23 billion words (2.97 billion tokens) and was split into training and validation sets at a 90:10 ratio for masked language modelling (MLM) pretraining. This broad coverage of water treatment terminology enabled WaterBERT to learn domain-specific language patterns during continual pretraining.

\subsection{Domain-adaptive pretraining of WaterBERT}\label{sec:dapt}

In this study, WaterBERT was developed through DAPT rather than pretraining a new model from scratch. Compared with pretraining from scratch, DAPT adapts a base language model using unlabeled domain-specific text, improving domain-specific semantic representations while preserving general linguistic knowledge in a more computationally efficient manner.\cite{ref07} Accordingly, we selected SciBERT as the base model for DAPT, because its scientific pretraining background is more consistent with water treatment literature than general-domain BERT.\cite{ref24}

Specifically, the vocabulary constructed from our water treatment corpus showed the highest meaningful overlap with SciBERT (57.75\%) among the evaluated BERT-based models, compared with 40.99\% for BERT-base (Table~\ref{tab:si2}). This suggests that SciBERT provides better lexical coverage of water treatment-related scientific text.

DAPT was performed on the constructed corpus using the MLM objective, starting from the SciBERT checkpoint. Next Sentence Prediction (NSP) was not included because the adaptation targeted token-level contextual representations, and prior BERT-family ablations found no consistent benefit from retaining the NSP loss.\cite{ref25} In MLM, a portion of the input tokens is randomly masked, and the model is trained to recover the original tokens from their surrounding context (Fig.~\ref{fig:workflow}).\cite{ref07} Whole-word masking was further adopted, whereby all subword tokens belonging to the same word are masked together, ensuring that complete domain-specific terms are treated as unified masking units. The masking probability was set to 0.15. The model was trained with an effective batch size of 160 and a peak learning rate of \(1\times10^{-4}\) using the AdamW optimizer.\cite{ref26} Training ran for 940,440 steps on a single NVIDIA A100 GPU, taking approximately 180 h, with the final checkpoint selected based on the lowest validation loss. All remaining hyperparameters are provided in Table~\ref{tab:si3}. Pseudo-perplexity (PPPL) was subsequently computed on the validation set for water treatment text (see \hyperref[text:si1]{Text S1} for details).\cite{ref27}

To further assess whether WaterBERT acquired stronger semantic representations of water treatment terminology through DAPT, a domain-specific CLOZE test was constructed. Two hundred water treatment-related articles published in \emph{Water Research} after February 2026 were selected, from which words with strong domain relevance, including mechanism, microbial, process, and operation, were masked to form fill-in-the-blank questions. All questions were manually reviewed to remove items without clear water treatment-specific meaning, yielding a final set of 230 CLOZE questions. Performance on the CLOZE test was evaluated using Acc@1, Acc@5, and Mean Reciprocal Rank (MRR), measuring whether the correct term was ranked first, within the top five predictions, and the average reciprocal rank of the correct answer, respectively.

\subsection{Downstream task annotation and fine-tuning}\label{sec:downstream}

To further assess whether the semantic advantages gained through DAPT translate into practical improvements in downstream water treatment text tasks, we fine-tuned and systematically evaluated WaterBERT on three labeled tasks: text classification, NER, and RE.

For the text classification task, we designed a multiclass classification task based on article abstracts to evaluate domain-specific semantic understanding. Using the Citation Topics framework from Web of Science, we selected five micro-topics closely related to wastewater treatment processes: activated sludge, adsorption, advanced oxidation, constructed wetland, and nanofiltration.

NER and RE were used to evaluate WaterBERT's ability to support structured information extraction from water treatment literature. A total of 1,000 abstracts were manually annotated for both tasks. For NER, eight entity types were defined: pollutant, water treatment process, reactor, treatment parameter, removal efficiency, value, microorganism, and dosed material, yielding 34,902 entities. To capture relationships among these entities, we further defined five RE types: ``targets'', ``has\_efficiency'', ``indicates\_removal\_of'', ``conditioned\_by'', and ``has\_value'', covering links among treatment processes, pollutants, removal performance, operating parameters, and corresponding values. A total of 14,471 relations were annotated for RE evaluation. Annotation dataset distributions are summarized in Tables~\ref{tab:si4} and~\ref{tab:si5}.

WaterBERT's performance on downstream tasks was benchmarked against six widely used BERT-based models representing different pretraining backgrounds: BERT-base\cite{ref16} and RoBERTa-base\cite{ref25} for general-domain text, SciBERT\cite{ref24} for scientific literature, and BioBERT\cite{ref28}, ClimateBERT\cite{ref29}, and EnvironmentalBERT\cite{ref30} for domain-specific corpora relevant to water treatment research. All models were trained and evaluated on the same dataset using five-fold cross-validation, with precision, recall, and F1 score reported as evaluation metrics.

\subsection{Large-scale corpus applications of WaterBERT}\label{sec:applications}

To demonstrate WaterBERT's capacity for large-scale text processing, we applied it to two corpus-level tasks: topic modelling and structured information extraction across the water treatment literature. WaterBERT was applied to topic modelling of 5,144 \emph{Environmental Science \& Technology (ES\&T)} articles published between 1995 and 2025 and extracted from the water treatment literature corpus. Article titles and abstracts were used as input. WaterBERT was used to generate document embeddings, followed by Uniform Manifold Approximation and Projection (UMAP) dimensionality reduction and Hierarchical Density-Based Spatial Clustering of Applications with Noise (HDBSCAN) clustering within BERTopic.\cite{ref31} Candidate parameter settings were screened according to topic granularity, noise proportion, cluster size balance, and interpretability, resulting in a final 35-topic model. The suitability of WaterBERT embeddings was further evaluated against SciBERT under matched BERTopic settings and multiple topic granularities using topic coherence, topic diversity, and clustering noise rate. Detailed procedures for BERTopic model selection and embedding evaluation are provided in \hyperref[text:si2]{Text S2}.

Using WaterBERT-fine-tuned models, we extracted structured knowledge from the corpus of 693,211 article abstracts to construct a water-domain knowledge graph. The NER model was first applied to all 693,211 abstracts to identify relevant entities. In parallel, a WaterBERT-based binary classifier was used to identify studies specifically related to pollutant removal, retaining 237,110 publications. For the binary-classification criteria, see \hyperref[text:si3]{Text S3}. RE was then performed only on this pollutant-removal subset, using the previously identified entities to extract treatment process--pollutant--removal efficiency records. Only records for which both NER and RE prediction confidence exceeded 0.90 were retained. After model inference, predicted entities were merged to consolidate equivalent mentions across the corpus. Details of the entity-merging procedure are provided in \hyperref[text:si4]{Text S4}. A subset of 300 abstracts was manually evaluated to assess extraction accuracy and benchmark performance against a range of LLMs.

Building on the resulting knowledge graph, we developed a Water Knowledge-Enhanced Retrieval System (WaterKERS) to facilitate retrieval and use of the extracted knowledge. A benchmark of 100 retrieval queries focusing on pollutant removal was used to evaluate retrieval performance. Two text-based retrieval methods, BM25-based lexical retrieval\cite{ref32} and embedding-based retrieval using BGE (BAAI/bge-large-en-v1.5),\cite{ref33} were used as baselines for comparison. WaterKERS was constructed by integrating knowledge graph-based retrieval with text-based retrieval. The detailed retrieval workflow of WaterKERS is described in \hyperref[text:si5]{Text S5} and illustrated in Fig.~\ref{fig:si3}. Retrieval relevance was evaluated using an LLM-as-a-judge framework, in which each retrieved document was classified as fully relevant, partially relevant, or irrelevant to the corresponding query. Retrieval performance was quantified using a relevance score based on these relevance categories. Details of the relevance criteria and evaluation procedure are provided in \hyperref[text:si6]{Text S6}.

\section{Results and discussion}\label{sec:results}

\subsection{Domain adaptation improves semantic understanding of water treatment text}\label{sec:domain-adaptation-results}

WaterBERT was designed to adapt SciBERT to wastewater treatment literature through continued masked language model (MLM) pretraining on a water treatment corpus. As shown in Fig.~\ref{fig:si1}, both the training and validation losses decrease consistently throughout pretraining. The training loss drops rapidly at the early stage and then declines more gradually, while the validation loss follows a similar downward trend, decreasing from 1.28 to nearly 1.02 without noticeable rebound, indicating that the model converged stably. After SciBERT was continually pretrained on the wastewater treatment corpus for approximately 1.0 million steps, the resulting WaterBERT achieved a PPPL of 2.33 on the validation set, substantially lower than 2.97 obtained by the original SciBERT. PPPL reflects how difficult it is for a model to predict masked words from their surrounding context. WaterBERT's lower PPPL means that it can predict missing words in wastewater-domain sentences more easily than SciBERT. These results provide intrinsic evidence that DAPT improved modelling of the language patterns in the water-treatment corpus.

\begin{figure}[t]
\centering
\includegraphics[width=0.98\linewidth]{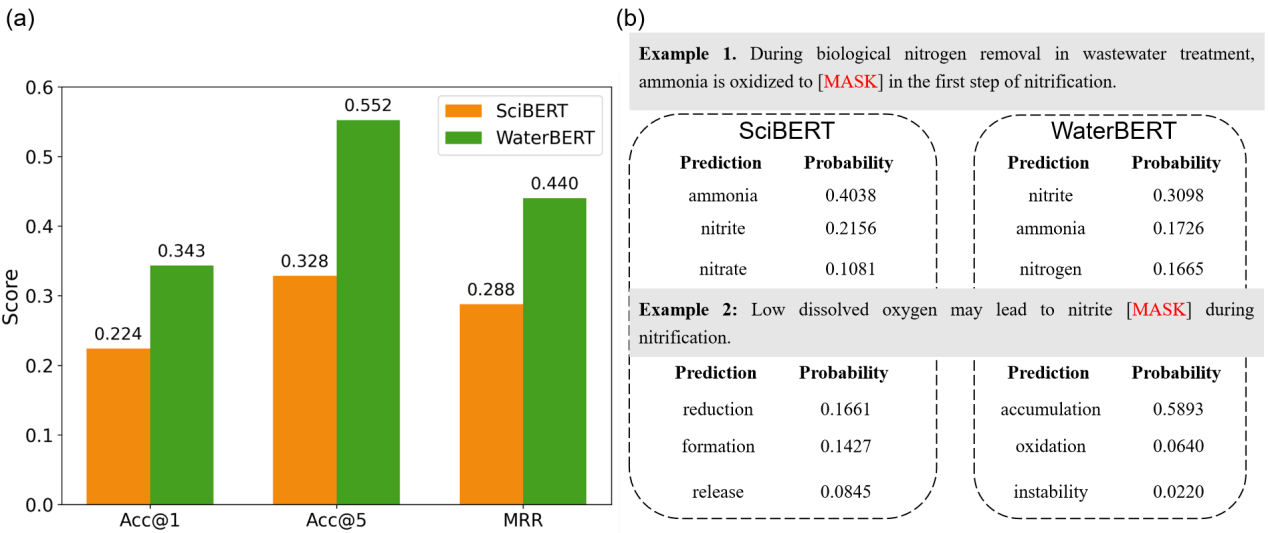}
\caption{Model performance on the CLOZE test was evaluated using three metrics (a): Acc@1 (the proportion of questions where the correct term was ranked as the top-1 prediction), Acc@5 (the proportion of questions where the correct term appeared within the top-5 predictions), and Mean Reciprocal Rank (MRR, the average of the reciprocal ranks of the correct answer across all questions); Representative CLOZE examples (b).}
\label{fig:cloze}
\end{figure}

We further evaluated the models using the CLOZE benchmark to examine whether WaterBERT had acquired stronger water treatment-specific semantic representations. In this benchmark, domain-relevant terms were masked in water treatment sentences, and the model was required to recover the missing term from its surrounding context. As shown in Fig.~\ref{fig:cloze}a, WaterBERT consistently outperformed SciBERT across all three CLOZE metrics. Acc@1 increased from 0.224 to 0.343, indicating that WaterBERT more often selected the correct wastewater-specific term as its top prediction. Acc@5 increased from 0.328 to 0.552, and MRR increased from 0.288 to 0.440, showing that WaterBERT also ranked correct terms higher among candidate predictions. These improvements suggest that continual pretraining on wastewater-domain corpora strengthened WaterBERT's ability to recover domain-relevant terms from treatment-specific contexts.

The representative examples in Fig.~\ref{fig:cloze}b offer a more intuitive illustration of WaterBERT's improved performance. In example 1, ``\emph{ammonia is oxidized to {[}MASK{]} in the first step of nitrification},'' the masked term refers to the product of ammonia oxidation, for which the correct answer is nitrite. WaterBERT correctly ranks ``nitrite'' first, indicating that it captures the reaction direction of nitrification and the product role of nitrite. By contrast, SciBERT ranks ``ammonia'' first, suggesting that it is more strongly influenced by the local co-occurrence between ``ammonia'' and ``nitrification'' but fails to infer the correct transformation relationship.\cite{ref34} Similarly, in example 2 ``\emph{Low dissolved oxygen may lead to nitrite {[}MASK{]} during nitrification,}'' the expected answer is accumulation, because oxygen limitation can inhibit the subsequent oxidation of nitrite to nitrate, causing nitrite production to exceed nitrite consumption.\cite{ref34} WaterBERT assigns a much higher probability to ``accumulation'' (0.58) than to other candidates, whereas SciBERT ranks the semantically opposite term ``reduction'' first and also predicts more generic process-related terms such as ``formation'' and ``release''. This contrast indicates that WaterBERT better captures the underlying mechanistic relationships in wastewater treatment processes, rather than simply selecting surface-level terms that appear semantically plausible. Overall, these results demonstrate that domain-adaptive pretraining enables WaterBERT to move beyond general scientific word associations and develop a stronger understanding of wastewater-specific process semantics.

\subsection{Benchmarking downstream water treatment text tasks}\label{sec:benchmarking}

To assess WaterBERT's performance in classifying water treatment abstracts, we compared its F1 score with those of the baseline models. As shown in Table~\ref{tab:si6}, WaterBERT achieved the highest macro F1 score of \(90.12 \pm 1.97\%\), compared with \(88.86 \pm 1.77\%\) for SciBERT. WaterBERT also achieved the highest F1 score across the five treatment process categories (Table~\ref{tab:si7}), demonstrating its improved ability to distinguish semantically related water treatment processes.

Text classification can identify relevant studies, but it cannot provide the structured information required to analyse treatment evidence. We therefore further evaluated WaterBERT on NER and RE. Compared with its base model SciBERT, WaterBERT increased the NER F1 score from \(76.13 \pm 0.69\%\) to \(79.50 \pm 0.33\%\) and the RE F1 score from \(70.51 \pm 0.64\%\) to \(74.04 \pm 0.56\%\), achieving the best overall performance in both tasks (Table~\ref{tab:ner-re}). WaterBERT substantially outperformed the general-domain BERT-base, with gains of 7.90\% for NER and 8.93\% for RE, indicating that general language representations are insufficient to capture the entity roles and relations specific to water treatment literature. More importantly, WaterBERT outperformed other domain-adapted models relevant to water treatment research, including BioBERT\cite{ref28}, ClimateBERT\cite{ref29}, and EnvironmentalBERT\cite{ref30}. The gap was particularly evident for EnvironmentalBERT, which only achieved F1 scores of 70.84\% for NER and 64.64\% for RE. This comparison suggests that an environmentally related pretraining model is insufficient to capture the entity roles and process semantics needed for water treatment information extraction.

\begin{figure}[t]
\centering
\includegraphics[width=0.98\linewidth]{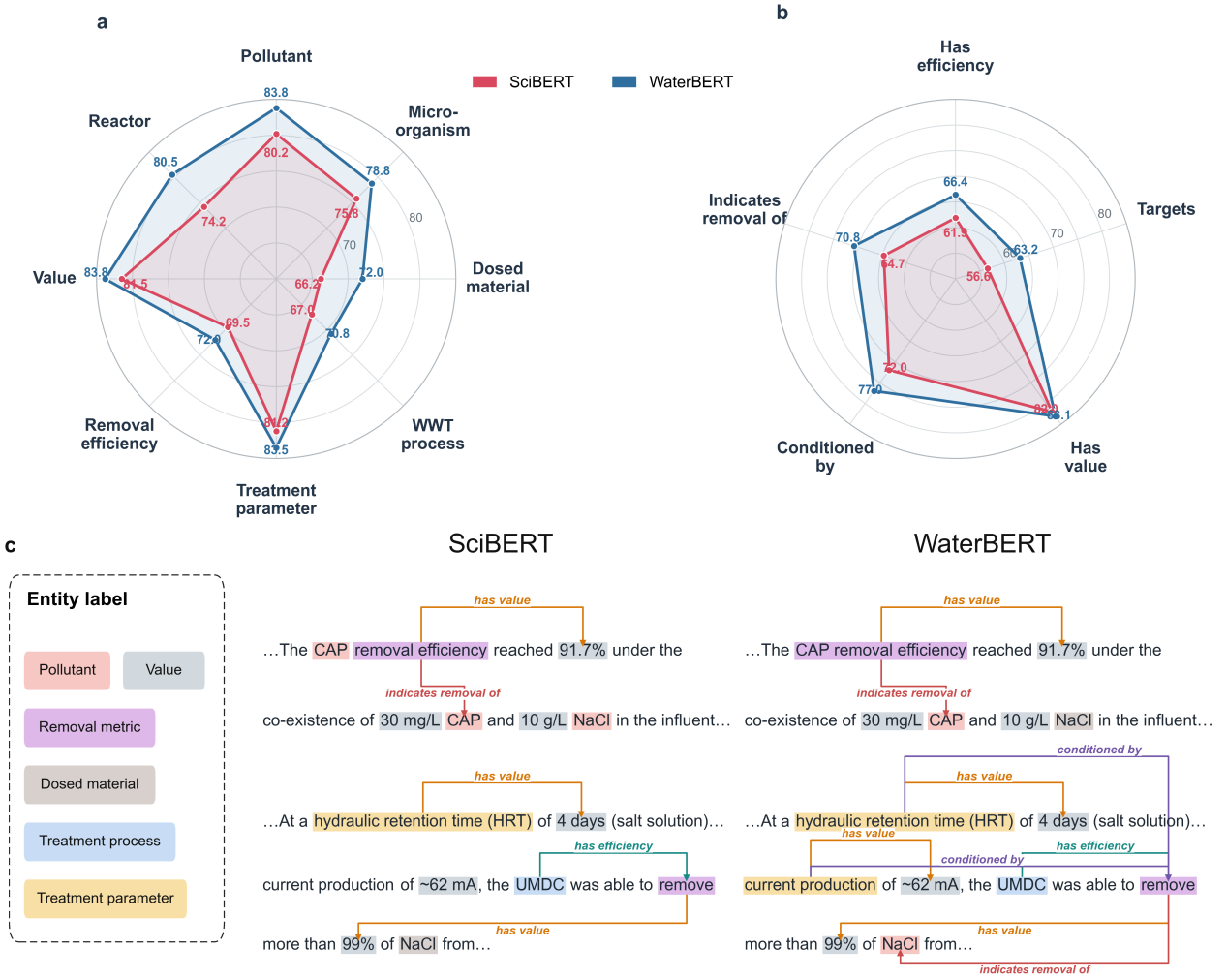}
\caption{Comparison of WaterBERT and SciBERT for information extraction in water treatment texts. (a) Performance on entity categories. (b) Performance on relation categories. (c) Visualization of entity extraction results on an example sentence.}
\label{fig:ie-comparison}
\end{figure}

The category-level results further clarify where these improvements originated (Fig.~\ref{fig:ie-comparison}a,b). WaterBERT showed larger gains over SciBERT for categories that require stronger interpretation of water treatment context. For NER, the improvements were greater for context-dependent entity types, including Dosed Material (+6.3\%), Reactor (+6.3\%), and Pollutant (+3.6\%), than for the more lexically explicit Value entity (+2.0\%). WaterBERT distinguished these context-dependent entity types more clearly, resulting in fewer missed entities and fewer cross-type misclassifications (Tables~\ref{tab:si8} and~\ref{tab:si9}). WaterBERT reduced false-negative rates across all entity types and reduced cross-type misclassifications from 356 to 233 (Fig.~\ref{fig:si2}). For example, Pollutant entities misclassified as Dosed Material decreased from 59 to 19, while Dosed Material entities misclassified as Pollutant decreased from 22 to 4. For RE, the improvements were likewise concentrated in semantically demanding relation types in water treatment contexts. F1 scores increased by 6.4\% for targets, 6.1\% for indicates\_removal\_of, and 5.0\% for conditioned\_by, compared with only 0.1\% for the more explicit has\_value relation.

The examples in Fig.~\ref{fig:ie-comparison}c illustrate why these context-dependent distinctions matter. NaCl is a typical example because its functional role varies across water treatment studies. It may appear as a background dosed material or electrolyte, such as in electrochemical systems, but can also be the target pollutant in studies of saline wastewater or salt removal.\cite{ref08,ref09} In the first example, NaCl is a coexisting background substance, but SciBERT incorrectly labels it as a Pollutant, whereas WaterBERT correctly identifies its contextual role. In the second example, SciBERT misclassifies NaCl as a Dosed Material rather than a Pollutant and misses the has\_value relation linking current production to approximately 62 mA. It consequently fails to recover the indicates\_removal\_of relation connecting the treatment process to NaCl and the conditioned\_by relation linking the process to its operating conditions. WaterBERT, by contrast, correctly identifies NaCl as a Pollutant and recovers the complete process--pollutant--condition structure. Overall, these results show that WaterBERT's improved domain-semantic representation translates into more reliable identification of treatment-specific entities and relations for structured information extraction.

\begin{table}[t]
\centering
\caption{Mean performance of NER and RE over five-fold cross-validation.}
\label{tab:ner-re}
\small
\setlength{\tabcolsep}{3.5pt}
\resizebox{\textwidth}{!}{%
\begin{tabular}{@{}lcccccc@{}}
\toprule
\textbf{Model} & \multicolumn{3}{c}{\textbf{NER}} & \multicolumn{3}{c}{\textbf{RE}} \\
\cmidrule(lr){2-4}\cmidrule(lr){5-7}
& \textbf{Precision} & \textbf{Recall} & \textbf{F1} & \textbf{Precision} & \textbf{Recall} & \textbf{F1} \\
\midrule
BERT-base & \(67.60\pm0.49\) & \(76.60\pm0.80\) & \(71.60\pm0.49\) & \(69.79\pm0.68\) & \(61.26\pm0.74\) & \(65.11\pm0.69\) \\
RoBERTa-base & \(71.60\pm0.49\) & \(80.00\pm0.63\) & \(75.80\pm0.40\) & \(68.83\pm0.65\) & \(66.93\pm0.68\) & \(67.89\pm0.64\) \\
BioBERT & \(72.20\pm0.40\) & \(80.40\pm1.02\) & \(75.80\pm0.75\) & \(72.32\pm0.62\) & \(66.45\pm0.70\) & \(70.17\pm0.66\) \\
ClimateBERT & \(68.48\pm0.69\) & \(77.32\pm1.03\) & \(72.63\pm0.60\) & \(65.58\pm0.72\) & \(71.95\pm0.65\) & \(68.62\pm0.68\) \\
EnvironmentalBERT & \(66.44\pm0.54\) & \(75.86\pm0.90\) & \(70.84\pm0.58\) & \(67.71\pm0.70\) & \(60.12\pm0.78\) & \(64.64\pm0.72\) \\
SciBERT & \(72.33\pm0.99\) & \(80.36\pm0.75\) & \(76.13\pm0.69\) & \(72.26\pm0.61\) & \(68.85\pm0.67\) & \(70.51\pm0.64\) \\
WaterBERT & \(76.50\pm0.53\) & \(82.75\pm0.48\) & \(79.50\pm0.33\) & \(72.65\pm0.58\) & \(75.49\pm0.60\) & \(74.04\pm0.56\) \\
\bottomrule
\end{tabular}%
}
\end{table}

\subsection{Topic modeling with WaterBERT--BERTopic}\label{sec:topic-modeling}

Topic modelling provides a systematic way to characterize thematic structures in water research. Conventional approaches, such as word frequency distributions and keyword co-occurrence analyses, rely largely on lexical patterns rather than contextual semantics and may therefore miss conceptually related studies expressed using different domain-specific terminology.\cite{ref35,ref36} BERTopic addresses this limitation through contextual embeddings, but its performance depends on how well the underlying BERT model represents the target domain. We therefore used WaterBERT, whose improved domain-semantic representation had been validated on water treatment texts, to generate water-domain document embeddings and applied BERTopic to 5,144 water treatment articles published in \emph{ES\&T}. We compared the resulting topic structure with that generated using SciBERT to evaluate whether domain adaptation improves topic coherence and interpretability.

As shown in Table~\ref{tab:topics}, WaterBERT outperformed SciBERT on all three metrics, with a higher normalized pointwise mutual information (NPMI) (0.117 vs. 0.077) and topic diversity (84.9\% vs. 80.7\%), and a lower noise ratio (24.4\% vs. 31.0\%). The higher NPMI and topic diversity indicate that WaterBERT-based topics exhibited stronger within-topic coherence and greater between-topic distinctiveness, respectively. These differences were also reflected in the top-10 topic representations, which showed clearer thematic separation and less overlap. In contrast, SciBERT showed greater repetition among the top-10 topic representations across different topics; for example, advanced oxidation- and Fenton-related terms repeatedly appeared in Topics 2, 5, and 10, indicating substantial thematic overlap among these topic clusters. WaterBERT also produced more semantically coherent and readily interpretable topic representations. For example, SciBERT's largest topic combined microbial fuel cells, ammonia-oxidizing bacteria, and activated sludge biofilms, despite these representing conceptually distinct research areas. In contrast, the corresponding WaterBERT topic (Topic 3) was more narrowly focused, with representative terms including ammonia oxidation, ammonium oxidation, nitrification, nitrifying, and nitrite, making it readily interpretable as biological nitrogen removal. Together, these results suggest that domain-adapted representations improve both topic separation and semantic coherence.

These improvements in topic separation and semantic coherence suggest that WaterBERT--BERTopic can serve as a practical approach for large-scale topic analysis in water treatment research. Compared with traditional tailored text mining pipelines, it reduces reliance on manually specified categories and lexical rules. For example, Zhu et al.~\cite{ref35} first defined research categories using domain knowledge and manually selected representative keywords for each category, and then identified research trends by tracking the frequencies of these predefined terms. Thus, the expansion of PFAS-related research was detected because terms such as PFAS, PFOA, and PFAA had been explicitly selected and monitored. In contrast, WaterBERT--BERTopic recovered a distinct PFAS-related theme (Topic 7) without specifying a PFAS category in advance. This is particularly useful for large-scale literature analysis, where manually defining keywords and categories becomes increasingly impractical and may constrain the identification of emerging or previously unrecognized research themes.

{\small
\setlength{\tabcolsep}{4pt}
\begin{longtable}{@{}>{\centering\arraybackslash}p{0.08\textwidth}p{0.42\textwidth}p{0.42\textwidth}@{}}
\caption{Comparison of WaterBERT and SciBERT topic models on the \emph{ES\&T} corpus}\label{tab:topics}\\
\toprule
\textbf{Metric} & \textbf{SciBERT} & \textbf{WaterBERT} \\
\midrule
\endfirsthead
\multicolumn{3}{r}{\tablename~\thetable\ (continued)}\\
\toprule
\textbf{Topic ID} & \textbf{SciBERT top 10 representation} & \textbf{WaterBERT top 10 representation} \\
\midrule
\endhead
\midrule
\multicolumn{3}{r}{Continued on next page}\\
\endfoot
\bottomrule
\endlastfoot
NPMI & 0.0769 & 0.1169 \\
Topic Diversity & 80.73\% & 84.94\% \\
Noise & 31.03\% & 24.42\% \\
\midrule
\textbf{Topic ID} & \multicolumn{2}{c}{\textbf{Top 10 representation}} \\
\midrule
1 & oxidizing bacteria; microbial fuel; microbial community; activated sludge; microbial communities; bacterial community; bacterial; ammonia oxidizing; biofilm; biofilms & sanitation; environmental impacts; environmental impact; sustainability; sustainable; biofuels; greenhouse gas; recycling; gas emissions; eutrophication \\
2 & electro fenton; electrocatalytic; photocatalytic activity; fe2o3; catalytic activity; photocatalytic; electrochemical; fenton like; fenton; photocatalysts & advanced oxidation; electro fenton; electrocatalytic; catalytic activity; electrochemical; catalytic; catalysis; h2o2; photocatalytic; fe2 \\
3 & sediments; contamination; watersheds; watershed; sediment; watershed scale; runoff; creek; wetlands; surface waters & ammonia oxidizing; ammonia oxidation; oxidizing bacteria; biological nitrogen; ammonium oxidation; nitrification denitrification; nitrification; nitrifying; nitrous acid; nitrite \\
4 & environmental sustainability; sustainability; environmental impacts; sustainable; environmental impact; consumption; emissions; greenhouse gas; gas emissions; resource recovery & chlorine disinfection; uv chlorine; disinfection byproducts; chlorine; chlorination; chloramination; chlorinated; disinfectants; ozone chlorine; free chlorine \\
5 & chlorine; chlorination; free chlorine; chloramines; hypochlorous acid; chlorinated; chloramine; uv h2o2; hydroxyl radical; advanced oxidation & contaminated groundwater; geochemical; aquifers; soils; sediments; soil; arsenic; acid drainage; phosphorus; minerals \\
6 & contaminated groundwater; minerals; leaching; phosphate; arsenic; soil; soils; contaminated; phosphorus; remediation & surface runoff; sediments; creek; sediment; runoff; watershed scale; watersheds; watershed; hydrologic; surface waters \\
7 & nanofiltration membranes; nanofiltration membrane; nanofiltration; pvdf membrane; nanofiltration nf; composite membranes; membranes; ultrafiltration; nf membranes; membrane fouling & polyfluoroalkyl substances; substances pfas; perfluorooctanoic acid; substances pfass; perfluorooctanoic; acid pfoa; perfluoroalkyl; polyfluoroalkyl; chemicals; sulfonic acid \\
8 & activated carbon; surfactants; hydrophobicity; hydrophobic; desorption; porous media; ionic; ionic strength; adsorbents; silicate & nanoparticles ag; silver nanoparticles; nanoparticles nps; nanoparticles; nanoparticle; sewage sludge; biosolids; microplastics; silver; nanomaterials \\
9 & estrogenic activity; estrogens; estrogen; estrogenic; hormone; androgen; fish exposed; rainbow trout; zebrafish; trout & nanofiltration; osmosis membranes; membrane fouling; composite membrane; membranes; ultrafiltration; membrane surface; ro membranes; reverse osmosis; ro membrane \\
10 & fenton reaction; advanced oxidation; oxidants; ii oxidation; oxidant; oxidized; peracetic acid; peroxymonosulfate; sulfate radical; catalyzed & iron oxides; As (iii) oxidation; Mn (ii) oxidation; ferrihydrite; arsenic; minerals; manganese oxide; corrosion; oxidation state; iron \\
\end{longtable}
}

\subsection{WaterBERT enables scalable and cost-effective knowledge construction from water treatment literature}\label{sec:knowledge-construction}

\begin{figure}[p]
\centering
\includegraphics[width=0.95\linewidth]{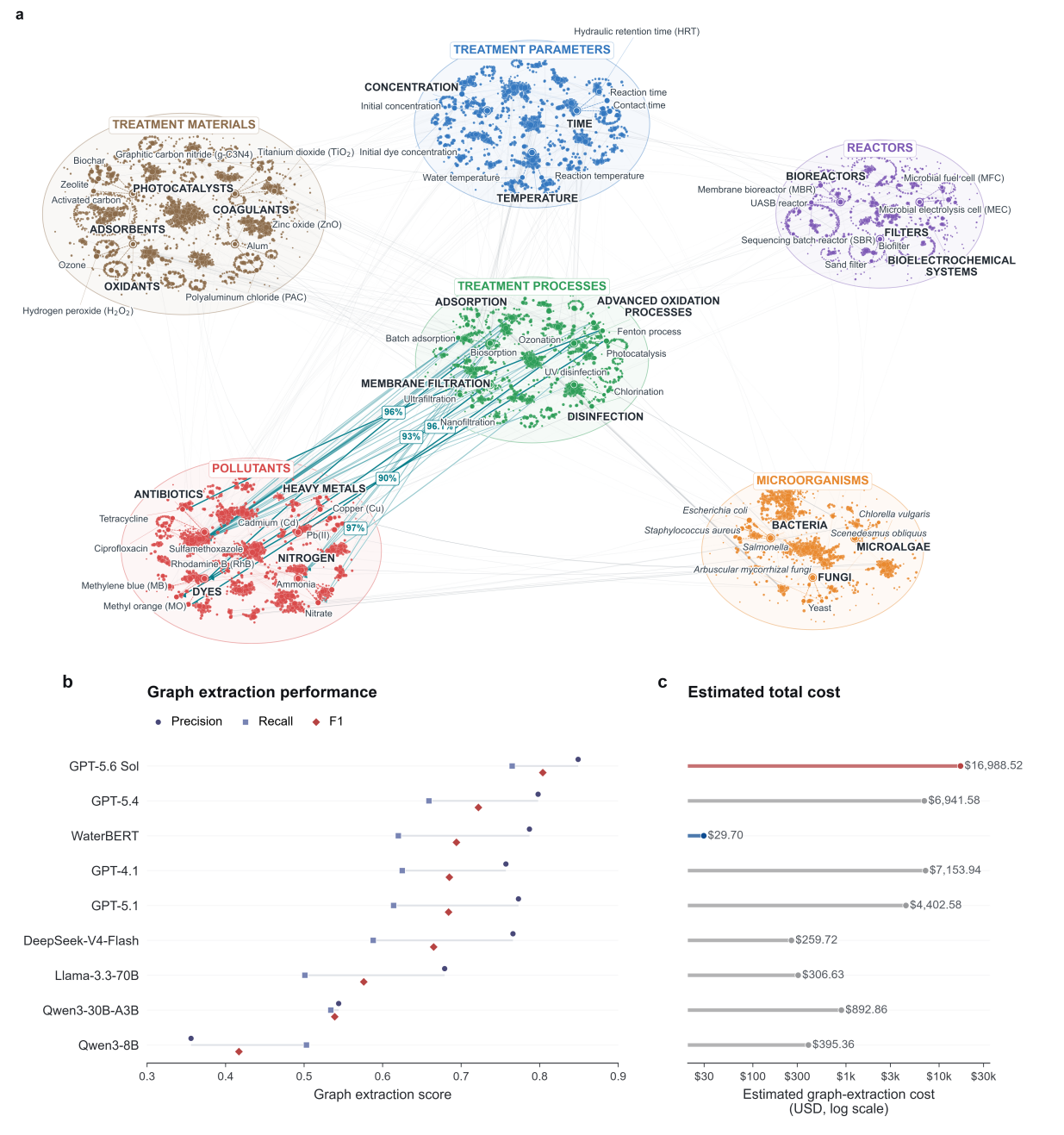}
\caption{Large-scale knowledge graph construction and extraction performance of WaterBERT. a, Representative subgraph extracted from the water treatment literature. Node colours denote six entity types: material, microorganism, treatment parameter, pollutant, treatment process, and reactor. Bold black labels denote higher-level concept nodes used to organize related entities. Green edges represent pollutant-removal relationships between treatment processes and pollutants, with reported removal efficiencies recorded as edge attributes. b, Precision, recall and F1 scores for graph extraction by WaterBERT and eight general-purpose language models, evaluated using 300 sampled articles. c, Estimated total extraction cost, in US dollars, for processing all 693,211 water treatment abstracts. Costs represent projected full-corpus deployment costs for each model.}
\label{fig:knowledge-graph}
\end{figure}

Evidence in the water treatment literature remains largely embedded in unstructured text. Materials, pollutants, treatment processes, operating parameters and removal outcomes are described using heterogeneous terminology, while the same term may assume different functional roles depending on context. Keyword- and embedding-based retrieval can identify documents with lexical overlap or broad semantic similarity, but cannot reliably organize these elements into connected, retrieval-ready facts. Knowledge graphs address this limitation by organizing entities and their relationships into interconnected, machine-queryable structures, enabling complex factual relationships and constraints to be represented explicitly.\cite{ref37,ref38} However, field-scale knowledge graphs for systematically structuring water treatment literature remain lacking. We therefore used our fine-tuned WaterBERT models for NER and RE as the extraction backbone to process 693,211 water treatment abstracts and construct a field-scale knowledge graph. As shown in Fig.~\ref{fig:knowledge-graph}a, the graph captures key information entities in water treatment, including materials, microorganisms, parameters, pollutants, treatment processes and reactors. Across the corpus, WaterBERT identified 8,800,205 retained entity mentions, which were consolidated into 118,131 canonical entities and connected by 4,249,707 edges across four relationship types: entity--publication occurrence, BELONGS\_TO, SUBCLASS\_OF, and REMOVES, providing a structured representation of evidence in the water treatment literature. Further details of entity consolidation and knowledge graph construction are provided in \hyperref[text:si4]{Text S4}.

To assess knowledge graph extraction reliability and cost, we manually evaluated 300 sampled articles and compared WaterBERT with frontier commercial LLMs and smaller open-weight LLMs. As shown in Fig.~\ref{fig:knowledge-graph}b, WaterBERT achieved a graph extraction F1 score of 69.3\%, lower than GPT-5.4 (72.2\%) and GPT-5.6 Sol (80.4\%) but substantially higher than all tested smaller open-weight models (57.6\% for Llama-3.3-70B, 53.9\% for Qwen3-30B-A3B and 41.7\% for Qwen3-8B). It also outperformed GPT-5.1 (68.4\%) and GPT-4.1 (68.5\%). Error analysis showed that smaller open-weight models more often confused entity boundaries and context-dependent functional roles (Table~\ref{tab:si10}). For example, they tended to label standalone concentration units, such as ppb and ppm, as Treatment\_Parameter entities or misclassify fixed treatment materials, such as ZIF-67-PAA, as reactors. In contrast, WaterBERT more accurately identified domain-specific spans and functional roles according to their local context. More importantly, WaterBERT approached the extraction accuracy of frontier commercial models at a substantially lower estimated deployment cost. For the full water treatment corpus of 693,211 abstracts, the projected extraction cost of the evaluated GPT models ranged from US\$4,402 to US\$16,988.52. By comparison, WaterBERT is an encoder-only model with only 110M parameters, and its estimated GPU rental cost for processing the same corpus was only US\$29.70 (Fig.~\ref{fig:knowledge-graph}c), while still achieving performance close to that of the frontier models such as GPT-4.1. This makes customized, corpus-scale knowledge graph construction feasible for water treatment research groups without extensive commercial computing resources.

\begin{figure}[t]
\centering
\includegraphics[width=0.98\linewidth]{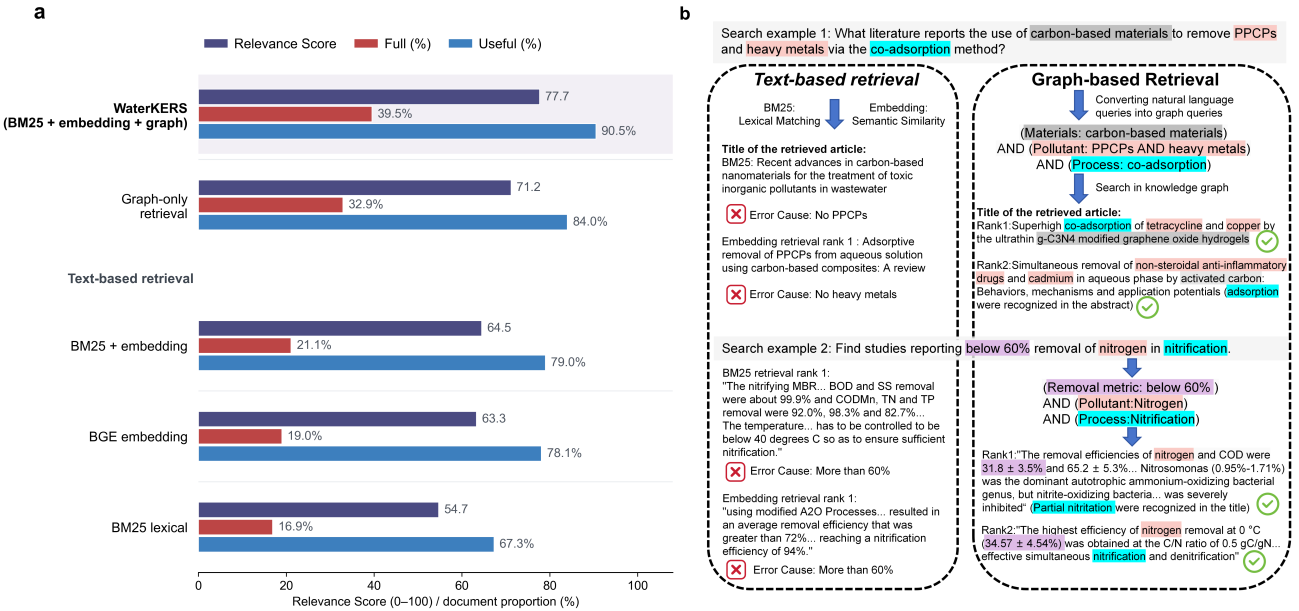}
\caption{Knowledge-graph-enhanced retrieval improves the relevance of retrieved literature. a, Retrieval performance of fusion retrieval, graph-only retrieval, hybrid BM25--embedding retrieval, BGE embedding retrieval and BM25. WaterKERS combines BM25 lexical matching, knowledge graph retrieval and BGE embeddings. Relevance Score measures overall retrieval relevance; b, Representative queries comparing conventional text-based retrieval with graph-based retrieval. Graph retrieval decomposes each query into typed entities, relationships and numerical constraints and requires these conditions to be jointly satisfied.}
\label{fig:retrieval}
\end{figure}

Building on this corpus-scale knowledge graph, we developed a graph-based retrieval approach and benchmarked it against text-based retrieval methods. As shown in Fig.~\ref{fig:retrieval}a, graph retrieval achieved a relevance score of 71.2, compared with 54.7--64.5 for the three text retrieval baselines: BM25 lexical matching, embedding retrieval, and their hybrid. Graph retrieval returned a higher proportion of fully relevant documents than text-based retrieval (32.9\% vs. 16.9--21.1\%). This improvement arose from the ability of graph retrieval to map query requirements onto structured facts and enforce explicit constraints on entities and their relationships, thereby retrieving documents that more precisely matched the query. However, for queries or constraints that could not be adequately represented by the graph schema, semantic text retrieval remained useful for identifying relevant documents (detailed analysis in \hyperref[text:si5]{Text S5}). We therefore integrated graph retrieval with text-based retrieval to construct a Water Knowledge-Enhanced Retrieval System (WaterKERS). This complementary strategy further increased the relevance score to 77.7 and raised the proportions of fully relevant and useful documents to 39.5\% and 90.5\%, respectively.

The examples in Fig.~\ref{fig:retrieval}b illustrate the sources of improvement enabled by graph-based retrieval. In the first example, the query asked for studies in which carbon-based materials were used to remove both pharmaceuticals and personal care products (PPCPs) and heavy metals through co-adsorption. BM25 and BGE retrieved studies on pollutant removal using carbon-based materials, but BM25 failed to satisfy the PPCP constraint, whereas BGE missed the heavy-metal constraint. Graph retrieval instead decomposed the query into typed constraints for the material, the two pollutant categories, and the treatment process, and required all four elements to co-occur. It therefore retrieved documents that satisfied the complete query. Notably, these documents did not explicitly use the umbrella term ``PPCPs'' but instead referred to specific compounds or subclasses, including antibiotics and nonsteroidal anti-inflammatory drugs. This terminology mismatch, combined with the requirement for multiple concepts to co-occur, caused text-based methods to rank these factually relevant documents outside the top 1,000 results. A similar pattern was observed in the second example, where the query requested studies reporting nitrogen removal efficiencies below 60\%. The text-based methods captured the general topic of nitrogen removal but could not reliably enforce the relationship between the removal metric and the numerical threshold, and therefore returned studies with efficiencies above 60\%. In contrast, graph retrieval explicitly linked the ``below 60\%'' constraint to nitrogen removal during nitrification and retrieved studies reporting efficiencies of \(31.8\pm3.5\%\) and \(34.57\pm4.54\%\).

\section{Implications}\label{sec:implications}

Extracting and analyzing information from the water treatment literature has long been essential to research in the field. Whether the goal is to develop data-driven models for contaminant removal,\cite{ref04,ref39} synthesize treatment performance across studies through meta-analysis,\cite{ref40,ref41} or trace the evolution of research topics,\cite{ref35,ref36} researchers have traditionally had to locate, screen, and interpret relevant information scattered across a vast body of publications. The emergence of LLMs offers a new way to automate these tasks, but their use involves a practical trade-off. Frontier proprietary models can interpret complex scientific text effectively but are costly to apply at corpus scale, whereas more affordable open-weight general models may struggle with the specialized terminology, experimental conditions, and treatment processes reported in water treatment studies.\cite{ref06,ref13,ref14} This creates a need for a scalable and affordable domain-specific language model for water treatment research.

WaterBERT addresses this need by providing a domain-specific semantic foundation that can be adapted to different water treatment literature-mining tasks. Through domain-adaptive pretraining, WaterBERT acquires stronger representations of treatment-specific terminology, contextual roles, and semantic relationships than its general scientific base model. Across downstream benchmarks, WaterBERT achieved the best overall performance among the evaluated BERT models, with F1 scores of 90.12\% for multiclass treatment process classification, 79.50\% for NER, and 74.04\% for RE. These capabilities make WaterBERT applicable across multiple stages of water treatment literature mining. For topic analysis, BERTopic based on WaterBERT embeddings can generate more interpretable water treatment subtopic clusters with less noise, without requiring predefined water treatment terms or categories such as pollutant classes, treatment processes, or manually specified keyword rules. For information extraction, WaterBERT can support a sequential workflow in which a classification model first screens publications according to relevant water treatment topics, followed by customized NER and RE models that extract domain-specific entities and their relationships.

More importantly, WaterBERT's compact 110M-parameter architecture and low inference cost make it particularly suitable for corpus-scale literature processing. Using WaterBERT-based models, we processed 693,211 abstracts to construct a field-scale knowledge graph containing 118,131 canonical entities and 4,249,707 relationships, at an estimated cost far lower than LLMs (US\$29.70 vs. US\$4,402--16,989). Building on this corpus-level knowledge graph, we developed WaterKERS and demonstrated how structured domain knowledge can support literature retrieval. As the first open-source corpus-level knowledge-graph retrieval system for water treatment research, WaterKERS provides an accessible platform for the water treatment community to organize and retrieve structured evidence based not only on semantic similarity but also on explicit factual knowledge. The broader value of this framework lies in its extensibility in both the scope of information extracted and its downstream use. WaterBERT can be further adapted to structure topic-specific evidence beyond the pollutant-removal knowledge represented in the current WaterKERS, including areas such as resource recovery, energy consumption, and carbon emissions. Its applications are also not limited to literature retrieval. For example, customized NER and RE models fine-tuned from WaterBERT could extract emission values, reactor configurations, operating conditions, and treatment performance from greenhouse gas studies in water treatment and link them into structured records for meta-analysis, data-driven modelling, and evidence synthesis.\cite{ref42} Given the large volume and disciplinary diversity of water treatment research, WaterBERT provides an effective and low-cost foundation for large-scale literature processing and the construction of structured knowledge bases tailored to different subdomains.

This work has several limitations. First, WaterBERT-based extraction relies on explicitly reported information in the literature; as a result, implicit or inconsistently described treatment processes, pollutants, operating conditions, and performance metrics may be missed or incorrectly linked. While high-confidence predictions enhance reliability, the resulting records are intended for corpus-scale organization rather than replacing expert-curated datasets. Second, the current pretraining corpus is dominated by scientific literature and may not fully capture engineering knowledge used in practice. Future research could incorporate sources such as textbooks or design manuals to strengthen representation of process design and operational terminology that is less frequently described in research articles. Third, although WaterKERS incorporates structured knowledge to address the limitations of general text-based retrieval, graph retrieval presents inherent constraints. Complex natural language queries do not always map cleanly onto graph structures, and retrieval is restricted to explicitly defined entities and relations. Therefore, developing embedding models specifically tailored to water treatment is necessary to better organize and retrieve knowledge across the rapidly expanding literature. Future work could focus on improving extraction accuracy, supporting continuous literature updates, and extending WaterBERT to additional water treatment tasks and datasets.

\section*{Data Availability}

All the models and frameworks involved in this study are available at:
\par\noindent\url{https://github.com/Mudi12138/WaterBERT}

\input{references}

%% file: supplementary.tex

\clearpage
\begingroup
\edef\SIoldfigure{\arabic{figure}}
\edef\SIoldtable{\arabic{table}}
\edef\SIoldequation{\arabic{equation}}
\renewcommand{\thefigure}{S\arabic{figure}}
\renewcommand{\thetable}{S\arabic{table}}
\renewcommand{\theequation}{S\arabic{equation}}
\makeatletter
\@ifpackageloaded{hyperref}{%
  \renewcommand*{\theHfigure}{si.\arabic{figure}}%
  \renewcommand*{\theHtable}{si.\arabic{table}}%
  \renewcommand*{\theHequation}{si.\arabic{equation}}%
}{}
\makeatother
\setcounter{figure}{0}
\setcounter{table}{0}
\setcounter{equation}{0}
\setlength{\LTcapwidth}{\linewidth}
\setlength{\emergencystretch}{2em}
\newcommand{\SIerr}[1]{{\setlength{\fboxsep}{1pt}\colorbox{yellow!25}{\strut #1}}}
\newcommand{\SIpm}[2]{\shortstack{#1\%\\$\pm$#2\%}}

\section*{Supporting Information}

\begin{center}
{\large\bfseries Domain-Adaptive Pretraining Enhances Water Treatment Semantic Representation for Large-Scale Structured Literature Mining\par}
\vspace{0.75em}
Mudi Zhai\textsuperscript{a}, Ruihong Qiu\textsuperscript{b},
Qingyun Zeng\textsuperscript{c,d}, T. David Waite\textsuperscript{a},
Bing-Jie Ni\textsuperscript{a}, Haoran Duan\textsuperscript{a,e,*}
\end{center}

\noindent\textit{\textsuperscript{a} UNSW Water Research Centre, School of Civil and Environmental Engineering, The University of New South Wales, Sydney, NSW 2052, Australia}

\noindent\textit{\textsuperscript{b} School of Electrical Engineering and Computer Science, The University of Queensland, Brisbane, QLD 4072, Australia}

\noindent\textit{\textsuperscript{c} Microsoft Copilot Studio AI, Redmond, WA 98052, United States}

\noindent\textit{\textsuperscript{d} Departments of Mathematics \& Department of Computer and Information Science, University of Pennsylvania, Philadelphia, PA 19104, United States}

\noindent\textit{\textsuperscript{e} Department of Civil Engineering, The University of Hong Kong, Pokfulam, Hong Kong SAR, China}

\noindent\textit{\textsuperscript{*}Corresponding Author: haoran.duan@hku.hk}

\subsection*{List of Texts}
\begin{description}
  \item[Text S1.] Calculation of pseudo-perplexity (PPPL)
  \item[Text S2.] BERTopic model selection and embedding evaluation
  \item[Text S3.] Binary classification criteria for pollutant removal
  \item[Text S4.] Steps for merging entities in knowledge graph
  \item[Text S5.] The retrieval process of Water Knowledge-Enhanced Retrieval System.
  \item[Text S6.] Evaluation of retrieval relevance and performance metrics
\end{description}

\subsection*{List of Figures}
\begin{description}
  \item[Figure S1.] Training and validation loss curves of WaterBERT during continued masked language model pretraining on the wastewater-treatment corpus.
  \item[Figure S2.] Confusion matrices for the five-fold cross-validation results of WaterBERT and SciBERT on named entity recognition.
  \item[Figure S3.] Workflow of the Water Knowledge-Enhanced Retrieval System (WaterKERS) integrating graph-based and semantic retrieval.
  \item[Figure S4.] Distribution and hierarchical composition of canonical entities in the WaterBERT knowledge graph.
\end{description}

\subsection*{List of Tables}
\begin{description}
  \item[Table S1.] Top 10 most frequent terms in the water-treatment corpus.
  \item[Table S2.] Comparison of vocabulary overlap and unique terms between pre-trained models and domain-specific text
  \item[Table S3.] WaterBERT training hyperparameters.
  \item[Table S4.] Distribution of named entity recognition data.
  \item[Table S5.] Distribution of relation extraction data.
  \item[Table S6.] Mean classification performance over five-fold cross-validation.
  \item[Table S7.] Per-class performance for multi-class classification across wastewater treatment categories over five-fold cross-validation.
  \item[Table S8.] Five-fold cross-validation named entity recognition error breakdown of SciBERT and WaterBERT by entity type.
  \item[Table S9.] Five-fold cross-validation relation extraction error breakdown of SciBERT and WaterBERT by relation type.
  \item[Table S10.] Examples of graph extraction from different models.
\end{description}

\subsection*{Text S1. Calculation of pseudo-perplexity (PPPL)}
\phantomsection\label{text:si1}

For a token sequence, pseudo-log-likelihood was calculated by replacing each evaluated token with [MASK] in turn, recording the log-probability assigned to the original token given all remaining tokens, and summing these values across evaluated positions. Pseudo-perplexity (PPPL) was obtained by exponentiating the negative mean log-probability; therefore, lower PPPL indicates better prediction of tokens. PPPL was calculated as:
\[
\operatorname{PLL}(x)=\sum_{i=1}^{N}\log p\!\left(x_i\mid x_{\setminus i}\right),
\qquad
\operatorname{PPPL}(x)=\exp\!\left[-\operatorname{PLL}(x)/N\right].
\]
where $x=(x_1,\ldots,x_N)$ denotes the input token sequence, $x_i$ is the original token at position $i$, $x_{\setminus i}$ denotes the sequence with $x_i$ replaced by [MASK], and $N$ is the total number of evaluated tokens. All tokens in each sequence were evaluated individually once.

\subsection*{Text S2. BERTopic model selection and embedding evaluation}
\phantomsection\label{text:si2}

The dataset comprised 5,144 \textit{Environmental Science \& Technology} articles published between 1995 and 2025 and extracted from the water treatment literature corpus. Article titles and available abstracts were concatenated, and WaterBERT embeddings were used as document representations for BERTopic.

For the final taxonomy, 384 combinations of UMAP and HDBSCAN parameters were screened, focusing on solutions containing approximately 25--45 topics. Candidate models were evaluated according to topic number, noise proportion, cluster-size balance, topic representations, and representative documents. The selected model used UMAP with 10 neighbours and 20 components and HDBSCAN with a minimum cluster size of 40, \texttt{min\_samples=1}, and a cluster-selection epsilon of 0.

WaterBERT was compared with SciBERT under controlled BERTopic settings in which only the document embeddings differed. Both models used identical documents, preprocessing, dimensionality-reduction and clustering parameters, topic-reduction procedures, and topic-representation models. The naturally detected HDBSCAN topics were reduced to multiple target topic granularities for comparison.

Topic coherence was measured using normalized pointwise mutual information (NPMI) among each topic's top 10 representation terms (Eq.~\ref{eq:si-npmi}). Term probabilities and joint probabilities were estimated from document-level occurrence in the evaluation corpus. Term pairs with no document-level co-occurrence were assigned an NPMI of $-1$. Pairwise values were averaged within each topic and subsequently across topics.
\begin{equation}
\operatorname{NPMI}(w_i,w_j)=
\frac{\ln\!\left[\dfrac{P(w_i,w_j)}{P(w_i)P(w_j)}\right]}
{-\ln P(w_i,w_j)}
\label{eq:si-npmi}
\end{equation}
\begin{equation}
\operatorname{TD}=
\frac{\left|\displaystyle\bigcup_{t=1}^{T}W_t\right|}
{\displaystyle\sum_{t=1}^{T}|W_t|}
\label{eq:si-td}
\end{equation}
\begin{equation}
R_{\mathrm{noise}}=\frac{N_{-1}}{N}
\label{eq:si-noise}
\end{equation}

Topic diversity was calculated as the proportion of unique terms across all top-10 topic representations (Eq.~\ref{eq:si-td}). The noise rate was defined as the proportion of documents assigned the HDBSCAN noise label, $-1$ (Eq.~\ref{eq:si-noise}). All metrics were calculated separately at each target topic granularity and then equally averaged.

\subsection*{Text S3. Binary classification criteria for pollutant removal}
\phantomsection\label{text:si3}

Abstracts were manually classified based on the information provided in their titles and abstracts. An abstract was labelled as positive only when it clearly described the removal of one or more pollutants by a specific treatment process. Both the treatment process and the target pollutant had to be explicitly identified, while the treatment medium could be either an aqueous phase, such as water or wastewater, or a sludge phase. Studies that focused solely on the occurrence, distribution, transport, transformation, fate, or environmental behaviour of pollutants without explicitly investigating their removal were classified as negative. In addition, pollutant removal had to be clearly stated as an objective or outcome of the study rather than inferred indirectly from the broader context (e.g., studies describing changes in pollutant concentrations during treatment without explicitly indicating that pollutant removal was investigated). Abstracts meeting these criteria were assigned to the positive class, whereas all others were assigned to the negative class.

\subsection*{Text S4. Steps for merging entities in knowledge graph}
\phantomsection\label{text:si4}

We consolidated synonymous entity mentions using a type-specific, frequency-aware normalization and clustering workflow. From 8,800,205 mentions extracted from 693,211 publications, we normalized case, whitespace, punctuation and parenthetical abbreviations, retaining normalized forms occurring at least five times. Candidate pairs were generated within each entity type using character 3-gram TF--IDF similarity, abbreviation matching and punctuation-insensitive alphanumeric keys, followed by frequency-ordered leader clustering. Clusters were evaluated using DeepSeek V4 Flash with representative source contexts under conservative instructions that permitted only strict synonym merging while preserving distinctions in hierarchy, chemical state, isotope, isomer, qualifiers and combined processes. The structured model outputs underwent deterministic validation and were subsequently manually reviewed before graph construction. The curated entities were projected back onto the complete corpus to construct a heterogeneous, multilayer knowledge graph comprising three principal node families---canonical entities, publications and hierarchical classification nodes---and relation types: entity--publication occurrence edges weighted by mention frequency, \texttt{BELONGS\_TO} edges linking entities to level-2 subclasses, \texttt{SUBCLASS\_OF} edges connecting level-2 subclasses to level-3 categories, and \texttt{REMOVES} connecting process to the removal of pollutant, with removal efficiency values attached as edge attributes. The entity layer contained 118,131 canonical entities organized into six domain classes: pollutants, wastewater-treatment processes, reactors, treatment parameters, microorganisms and dosed materials. The distributions of canonical entities and mentions across the major level-2 subclasses of these six domain classes are shown in Fig.~\ref{fig:si4}.

\subsection*{Text S5. The retrieval process of Water Knowledge-Enhanced Retrieval System.}
\phantomsection\label{text:si5}

To prevent retrieval failure when a query cannot be faithfully expressed as graph facts, WaterKERS fuses graph-based retrieval with text-based retrieval. Graph retrieval performs well when the entities, relationships, and constraints specified in a query can be mapped directly onto the knowledge graph, but its effectiveness decreases when the query contains information that is not represented by the graph schema. For example, for the query ``Which studies report above 90\% removal of oil-in-water emulsion using adsorbent-based treatment?'', the graph correctly mapped ``oil'' and ``adsorption'' to relevant entities and relationships and therefore retrieved many studies involving adsorption-based oil removal. However, the graph schema did not encode the directional distinction between oil-in-water and water-in-oil emulsions. Consequently, graph retrieval also returned studies on crude-oil sorption for oil-spill response and water-in-oil emulsion separation. Although these studies shared relevant entities and processes with the query, they did not satisfy its specific requirement for oil-in-water emulsion treatment and were therefore only partially relevant or irrelevant. This example illustrates that structured retrieval can enforce constraints only when the required distinctions are explicitly represented in the graph. In such cases, relying on graph retrieval alone may return incomplete or misaligned results, whereas text-based retrieval can recover relevant documents by matching surface forms and broader semantic context.

The fusion pipeline works as follows (Fig.~\ref{fig:si3}). A first large language model (LLM1) translates the natural-language query into an executable query plan, decomposing it into atomic entity conditions and grouping them by logical intent: conditions that must co-occur are placed in separate groups joined by AND, while interchangeable alternatives are grouped within an OR. A second LLM (LLM2) then maps each entity condition onto the graph's classification hierarchy (L3/L2), so that a broad concept such as ``carbon-based materials'' resolves to its category, ``carbon-based adsorbents,'' which the graph expands to all papers in that category. Candidate papers are obtained by intersecting the paper sets of all AND-joined groups. Within the graph candidate pool, papers are ranked by BM25, a classic term-weighted text-scoring algorithm, and the top five are retained. In parallel, the query is embedded and matched against a dense vector index (BGE) to retrieve the semantically similar papers by cosine similarity. The final result fuses the two: the graph's top five form the core, and non-overlapping vector-retrieved papers fill the remaining five slots, yielding a final top-ten ranking that inherits the graph's precision when facts hold and the vector index's semantic coverage when they do not.

\subsection*{Text S6. Evaluation of retrieval relevance and performance metrics}
\phantomsection\label{text:si6}

We evaluated retrieval quality using a benchmark of 100 queries. Each query was issued to five retrieval systems---WaterKERS (graph--vector fusion), pure graph retrieval, hybrid BM25--BGE retrieval, dense embedding retrieval (BGE), and lexical retrieval (BM25)---with each system returning a ranked list of up to 10 candidate documents from a corpus of 693,211 wastewater-treatment papers. Relevance was judged by a large language model (GPT-5.6-sol), which received each query together with the titles and abstracts of the retrieved documents in a single prompt and assigned each document one of three labels with a one-sentence justification. A document was considered fully relevant if it satisfied all core information and relationships specified in the query; partially relevant if it addressed the core information but omitted or differed in one or more query constraints; and not relevant if it failed to address the core information, mentioned the topic only incidentally, or did not satisfy explicitly specified numerical constraints. For each query, retrieval relevance was quantified using a weighted relevance score (RS):
\[
\operatorname{RS}=\frac{N_{\mathrm{fully}}+0.75\times N_{\mathrm{partial}}}{K}\times100.
\]
where $N_{\mathrm{fully}}$ and $N_{\mathrm{partial}}$ denote the numbers of fully relevant and partially relevant documents, respectively, and $K$ is the total number of retrieved documents evaluated for the query.

All retrieval labels across all models were evaluated three times. The consistency of the LLM-as-a-judge approach reached 91.37\% (Cohen's kappa = 0.854), and any inconsistent results across the three runs were manually adjudicated.

\clearpage
\begin{figure}[p]
  \centering
  \includegraphics[width=\linewidth,height=0.78\textheight,keepaspectratio]{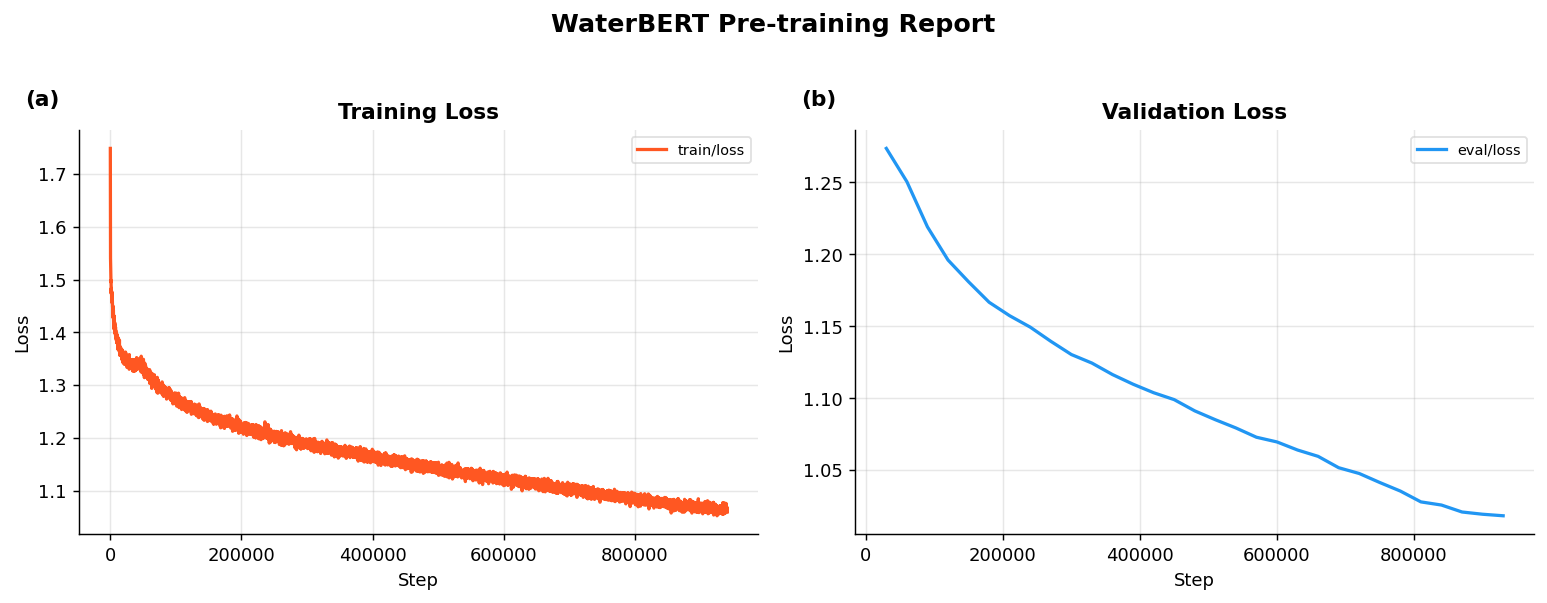}
  \caption{Training and validation loss curves of WaterBERT during continued masked language model pretraining on the wastewater-treatment corpus.}
  \label{fig:si1}
\end{figure}

\begin{figure}[p]
  \centering
  \includegraphics[width=\linewidth,height=0.78\textheight,keepaspectratio]{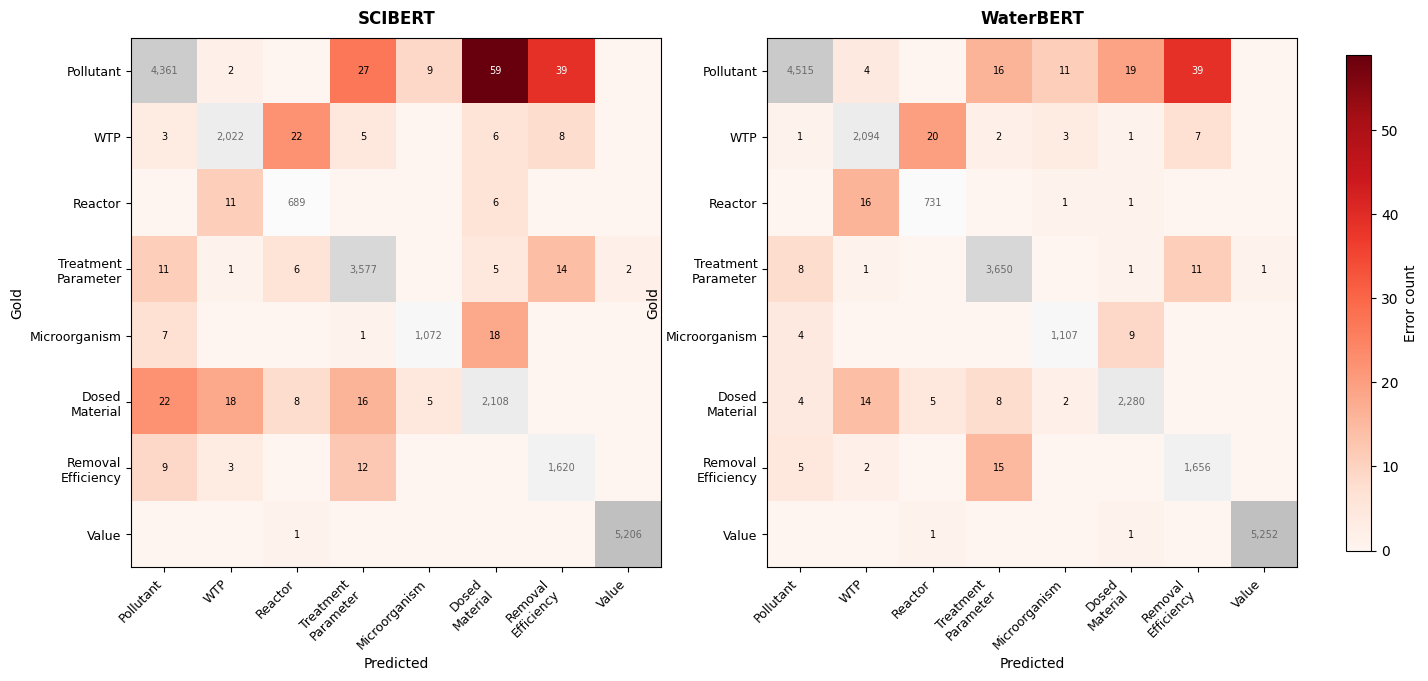}
  \caption{Confusion matrices for the five-fold cross-validation results of WaterBERT and SciBERT on named entity recognition.}
  \label{fig:si2}
\end{figure}

\begin{figure}[p]
  \centering
  \includegraphics[width=\linewidth,height=0.82\textheight,keepaspectratio]{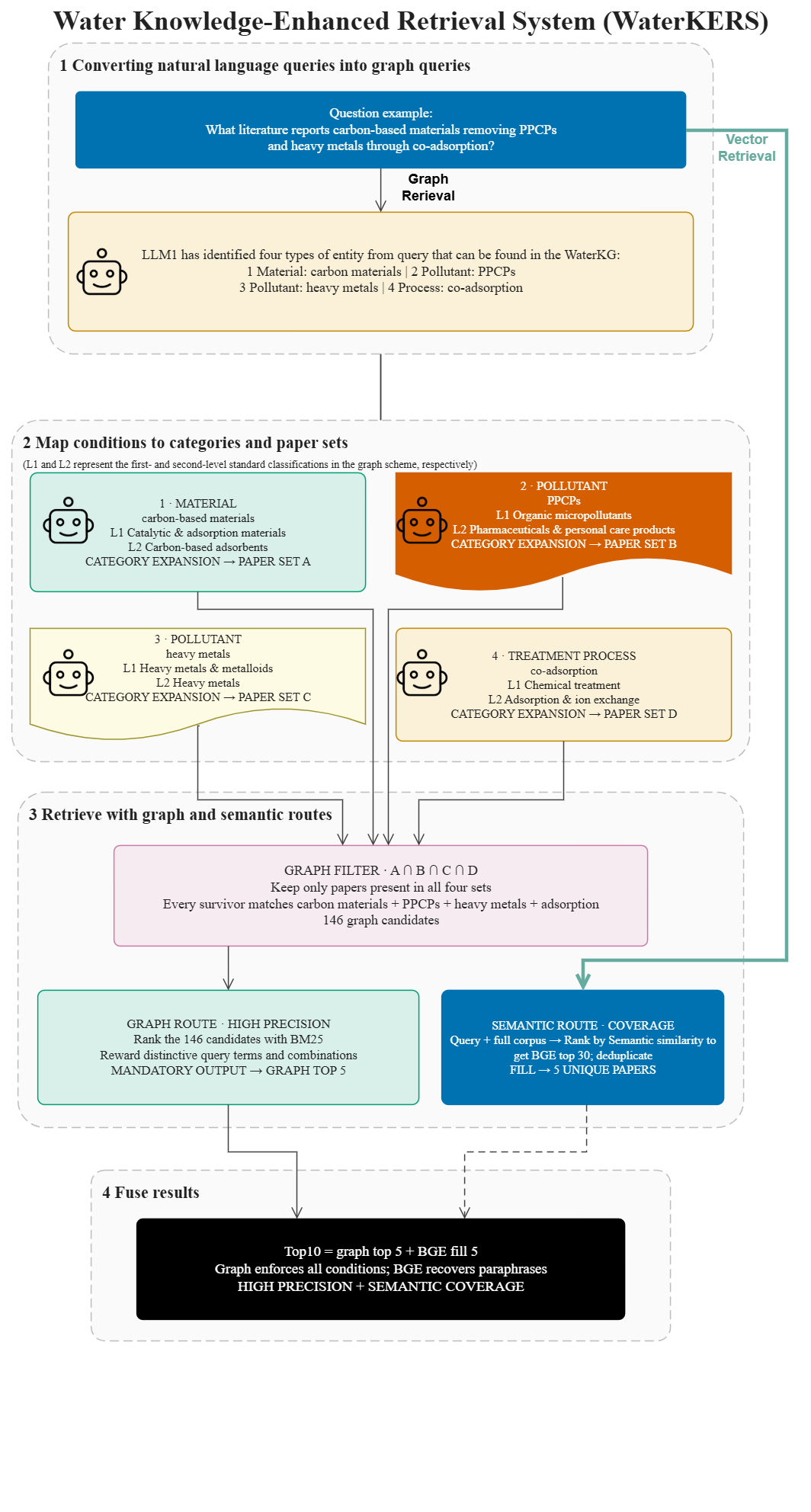}
  \caption{Workflow of the Water Knowledge-Enhanced Retrieval System (WaterKERS) integrating graph-based and semantic retrieval.}
  \label{fig:si3}
\end{figure}

\begin{figure}[p]
  \centering
  \includegraphics[width=\linewidth,height=0.82\textheight,keepaspectratio]{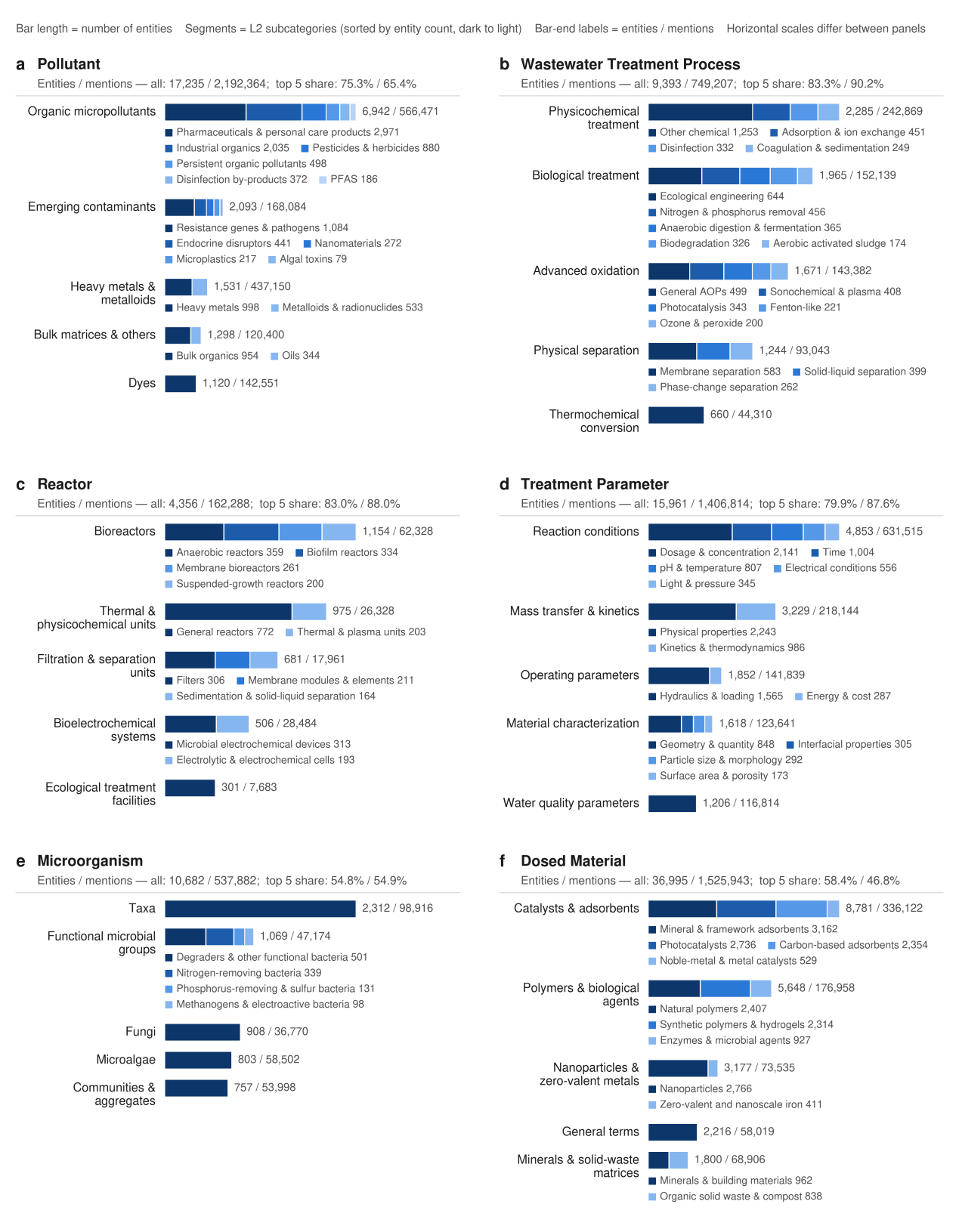}
  \caption{Distribution and hierarchical composition of canonical entities in the WaterBERT knowledge graph.}
  \label{fig:si4}
\end{figure}
\clearpage

\small
\begin{longtable}{@{}ccl@{}}
\caption{Top 10 most frequent terms in the water-treatment corpus.}\label{tab:si1}\\
\toprule
\textbf{Rank} & \textbf{Term} & \textbf{Frequency} \\
\midrule
\endfirsthead
\toprule
\textbf{Rank} & \textbf{Term} & \textbf{Frequency} \\
\midrule
\endhead
\bottomrule
\endlastfoot
1 & water & 8,787,412 \\
2 & treatment & 5,177,063 \\
3 & surface & 3,880,860 \\
4 & removal & 3,569,268 \\
5 & concentration & 3,440,378 \\
6 & adsorption & 3,435,441 \\
7 & wastewater & 3,324,296 \\
8 & process & 3,141,992 \\
9 & pH & 3,112,520 \\
10 & solution & 2,519,577 \\
\end{longtable}

\begin{longtable}{@{}lccc@{}}
\caption{Comparison of vocabulary overlap and unique terms between pre-trained models and domain-specific text.}\label{tab:si2}\\
\toprule
\textbf{Model} & \textbf{Vocabulary Size} & \textbf{Overlap} & \textbf{Custom Unique Words} \\
\midrule
\endfirsthead
\toprule
\textbf{Model} & \textbf{Vocabulary Size} & \textbf{Overlap} & \textbf{Custom Unique Words} \\
\midrule
\endhead
\bottomrule
\endlastfoot
SciBERT & 31,090 & 57.75 & 2,097 \\
ClimateBERT & 50,500 & 56.05 & 2,267 \\
RoBERTa-base & 50,265 & 55.99 & 2,273 \\
EnvironmentalBERT & 50,265 & 55.99 & 2,273 \\
BERT-base & 30,522 & 40.99 & 3,773 \\
BioBERT & 28,996 & 37.42 & 4,130 \\
\end{longtable}

\begin{longtable}{@{}p{0.42\linewidth}p{0.52\linewidth}@{}}
\caption{WaterBERT training hyperparameters.}\label{tab:si3}\\
\toprule
\textbf{Parameter} & \textbf{Setting} \\
\midrule
\endfirsthead
\toprule
\textbf{Parameter} & \textbf{Setting} \\
\midrule
\endhead
\bottomrule
\endlastfoot
Masking probability & 0.15 \\
Training steps & 940,440 \\
Training time & $\sim$180 h \\
GPU & NVIDIA A100 \\
Gradient accumulation steps & 2 \\
Per-device batch size & 80 \\
Effective batch size & 160 \\
Learning rate & $1\times10^{-4}$ \\
Warmup ratio & 0.048 \\
Weight decay & $1\times10^{-2}$ \\
Adam $\beta_1/\beta_2$ & 0.9 / 0.98 \\
Adam $\varepsilon$ & $1\times10^{-6}$ \\
Random seed & 42 \\
Initialization checkpoint & \texttt{allenai/scibert\_scivocab\_uncased} \\
Model architecture & BERT-base encoder (12 layers; hidden size 768; 12 attention heads) \\
Trainable parameters & 109,951,090 ($\sim$110 million) \\
Tokenizer and vocabulary & SciVocab uncased; 31,090 tokens \\
Maximum sequence length & 512 tokens \\
Pretraining objective & Masked language modelling \\
Masking strategy & Dynamic whole-word masking (WWM) \\
Token-replacement policy & 80\% [MASK]; 10\% random token; 10\% unchanged \\
Training epochs & 30 \\
Learning-rate scheduler & Linear \\
Maximum gradient norm & 1.0 \\
Training precision & bfloat16; TF32 enabled \\
Model-selection metric & Minimum validation loss \\
Transformers version & 4.57.3 (checkpoint metadata) \\
\end{longtable}

\begin{longtable}{@{}>{\raggedright\arraybackslash}p{0.20\linewidth}>{\centering\arraybackslash}p{0.20\linewidth}>{\centering\arraybackslash}p{0.14\linewidth}>{\centering\arraybackslash}p{0.30\linewidth}@{}}
\caption{Distribution of named entity recognition data.}\label{tab:si4}\\
\toprule
\textbf{Entity type} & \textbf{Number of labeled entities} & \textbf{Percentage (\%)} & \textbf{Number of articles containing the entity} \\
\midrule
\endfirsthead
\toprule
\textbf{Entity type} & \textbf{Number of labeled entities} & \textbf{Percentage (\%)} & \textbf{Number of articles containing the entity} \\
\midrule
\endhead
\bottomrule
\endlastfoot
Pollutant & 7614 & 21.8 & 947 \\
Wastewater Treatment Process & 4053 & 11.6 & 874 \\
Reactor & 1257 & 3.6 & 330 \\
Treatment Parameter & 5602 & 16.1 & 951 \\
Microorganism & 1725 & 4.9 & 298 \\
Dosed Material & 4308 & 12.3 & 681 \\
Removal Efficiency & 2811 & 8.1 & 945 \\
Value & 7532 & 21.6 & 972 \\
\end{longtable}

\begin{longtable}{@{}>{\raggedright\arraybackslash}p{0.24\linewidth}>{\centering\arraybackslash}p{0.18\linewidth}>{\centering\arraybackslash}p{0.14\linewidth}>{\centering\arraybackslash}p{0.28\linewidth}@{}}
\caption{Distribution of relation extraction data.}\label{tab:si5}\\
\toprule
\textbf{Relation type} & \textbf{Number of labeled relations} & \textbf{Percentage (\%)} & \textbf{Number of articles containing the relations} \\
\midrule
\endfirsthead
\toprule
\textbf{Relation type} & \textbf{Number of labeled relations} & \textbf{Percentage (\%)} & \textbf{Number of articles containing the relations} \\
\midrule
\endhead
\bottomrule
\endlastfoot
\texttt{targets} & 2769 & 19.1 & 809 \\
\texttt{has\_efficiency} & 1796 & 12.4 & 812 \\
\texttt{indicates\_removal\_of} & 2308 & 16.0 & 785 \\
\texttt{conditioned\_by} & 2883 & 19.9 & 729 \\
\texttt{has\_value} & 4715 & 32.6 & 885 \\
\end{longtable}

\begin{longtable}{@{}lccccc@{}}
\caption{Mean classification performance over five-fold cross-validation.}\label{tab:si6}\\
\toprule
\textbf{Models} & \textbf{Precision} & \textbf{Recall} & \textbf{F1} & \textbf{$\Delta$F1 95\% CI} & \textbf{Holm-adj $p$} \\
\midrule
\endfirsthead
\toprule
\textbf{Models} & \textbf{Precision} & \textbf{Recall} & \textbf{F1} & \textbf{$\Delta$F1 95\% CI} & \textbf{Holm-adj $p$} \\
\midrule
\endhead
\bottomrule
\endlastfoot
BERT-base & $87.86\pm1.67$ & $87.80\pm1.69$ & $87.79\pm1.68$ & $[+1.5,+3.2]$ & 0.0003 *** \\
RoBERTa-base & $87.87\pm2.44$ & $87.80\pm2.44$ & $87.79\pm2.45$ & $[+1.5,+3.2]$ & 0.0003 *** \\
BioBERT & $88.78\pm1.85$ & $88.75\pm1.84$ & $88.74\pm1.85$ & $[+0.6,+2.2]$ & 0.0016 ** \\
ClimateBERT & $88.08\pm1.97$ & $88.06\pm1.96$ & $88.04\pm1.96$ & $[+1.3,+2.9]$ & 0.0005 *** \\
EnvironmentalBERT & $87.67\pm1.69$ & $87.63\pm1.68$ & $87.62\pm1.68$ & $[+1.7,+3.4]$ & 0.0002 *** \\
SciBERT & $88.91\pm1.76$ & $88.87\pm1.78$ & $88.86\pm1.77$ & $[+0.4,+2.1]$ & 0.0027 ** \\
WaterBERT & $90.16\pm1.99$ & $90.12\pm1.98$ & $90.12\pm1.97$ & / & / \\
\end{longtable}

\clearpage
\begin{landscape}
\scriptsize
\setlength{\tabcolsep}{1.5pt}
\renewcommand{\arraystretch}{1.2}
\begin{longtable}{@{}l*{15}{c}@{}}
\caption{Per-class performance for multi-class classification across wastewater treatment categories over five-fold cross-validation.}\label{tab:si7}\\
\toprule
\textbf{Model} & \multicolumn{3}{c}{\textbf{Activated Sludge}} & \multicolumn{3}{c}{\textbf{Adsorption}} & \multicolumn{3}{c}{\textbf{Advanced Oxidation}} & \multicolumn{3}{c}{\textbf{Constructed Wetland}} & \multicolumn{3}{c}{\textbf{Nanofiltration}} \\
& \textbf{Precision} & \textbf{Recall} & \textbf{F1} & \textbf{Precision} & \textbf{Recall} & \textbf{F1} & \textbf{Precision} & \textbf{Recall} & \textbf{F1} & \textbf{Precision} & \textbf{Recall} & \textbf{F1} & \textbf{Precision} & \textbf{Recall} & \textbf{F1} \\
\midrule
\endfirsthead
\toprule
\textbf{Model} & \multicolumn{3}{c}{\textbf{Activated Sludge}} & \multicolumn{3}{c}{\textbf{Adsorption}} & \multicolumn{3}{c}{\textbf{Advanced Oxidation}} & \multicolumn{3}{c}{\textbf{Constructed Wetland}} & \multicolumn{3}{c}{\textbf{Nanofiltration}} \\
& \textbf{Precision} & \textbf{Recall} & \textbf{F1} & \textbf{Precision} & \textbf{Recall} & \textbf{F1} & \textbf{Precision} & \textbf{Recall} & \textbf{F1} & \textbf{Precision} & \textbf{Recall} & \textbf{F1} & \textbf{Precision} & \textbf{Recall} & \textbf{F1} \\
\midrule
\endhead
\midrule
\multicolumn{16}{r}{Continued on next page}\\
\endfoot
\bottomrule
\endlastfoot
BERT-base & \SIpm{85.31}{3.65} & \SIpm{85.31}{1.83} & \SIpm{85.28}{2.26} & \SIpm{88.78}{3.36} & \SIpm{88.98}{2.55} & \SIpm{88.84}{2.10} & \SIpm{87.54}{1.45} & \SIpm{88.75}{3.87} & \SIpm{88.09}{1.79} & \SIpm{89.00}{1.60} & \SIpm{86.97}{3.23} & \SIpm{87.94}{1.72} & \SIpm{88.69}{2.71} & \SIpm{88.99}{2.93} & \SIpm{88.83}{2.63} \\
RoBERTa-base & \SIpm{85.09}{4.59} & \SIpm{85.78}{2.14} & \SIpm{85.41}{3.22} & \SIpm{88.53}{3.91} & \SIpm{88.75}{3.01} & \SIpm{88.55}{1.63} & \SIpm{88.17}{1.87} & \SIpm{87.45}{4.75} & \SIpm{87.76}{2.89} & \SIpm{87.66}{2.60} & \SIpm{89.22}{4.23} & \SIpm{88.41}{3.11} & \SIpm{89.88}{2.77} & \SIpm{87.80}{5.06} & \SIpm{88.81}{3.84} \\
BioBERT & \SIpm{85.59}{2.74} & \SIpm{85.78}{3.57} & \SIpm{85.67}{2.87} & \SIpm{90.37}{2.18} & \SIpm{90.40}{1.97} & \SIpm{90.35}{0.55} & \SIpm{88.10}{1.66} & \SIpm{89.46}{2.65} & \SIpm{88.76}{1.97} & \SIpm{89.32}{3.65} & \SIpm{88.39}{2.30} & \SIpm{88.83}{2.50} & \SIpm{90.54}{2.14} & \SIpm{89.70}{4.08} & \SIpm{90.10}{2.90} \\
ClimateBERT & \SIpm{85.45}{3.76} & \SIpm{85.19}{2.47} & \SIpm{85.31}{3.00} & \SIpm{89.11}{2.88} & \SIpm{89.22}{2.22} & \SIpm{89.12}{1.37} & \SIpm{88.89}{2.12} & \SIpm{89.22}{4.66} & \SIpm{89.03}{3.09} & \SIpm{88.62}{2.05} & \SIpm{87.68}{3.47} & \SIpm{88.13}{2.53} & \SIpm{88.32}{2.34} & \SIpm{88.99}{4.59} & \SIpm{88.64}{3.38} \\
EnvironmentalBERT & \SIpm{85.00}{1.94} & \SIpm{84.71}{3.06} & \SIpm{84.85}{2.42} & \SIpm{88.48}{2.41} & \SIpm{89.22}{3.77} & \SIpm{88.78}{1.53} & \SIpm{89.79}{1.36} & \SIpm{87.56}{4.06} & \SIpm{88.63}{2.46} & \SIpm{87.55}{3.10} & \SIpm{87.68}{2.30} & \SIpm{87.58}{1.94} & \SIpm{87.52}{1.59} & \SIpm{88.99}{3.32} & \SIpm{88.23}{2.28} \\
SciBERT & \SIpm{85.62}{3.81} & \SIpm{85.19}{3.44} & \SIpm{85.40}{3.57} & \SIpm{90.90}{2.30} & \SIpm{90.17}{2.39} & \SIpm{90.49}{0.36} & \SIpm{89.48}{3.22} & \SIpm{89.34}{4.00} & \SIpm{89.38}{3.16} & \SIpm{89.95}{1.92} & \SIpm{88.75}{2.36} & \SIpm{89.32}{1.47} & \SIpm{88.61}{2.58} & \SIpm{90.88}{3.19} & \SIpm{89.70}{2.33} \\
WaterBERT & \SIpm{88.25}{2.36} & \SIpm{88.03}{3.44} & \SIpm{88.12}{2.58} & \SIpm{92.57}{1.30} & \SIpm{90.05}{3.05} & \SIpm{91.28}{1.99} & \SIpm{90.84}{2.29} & \SIpm{91.47}{2.23} & \SIpm{91.15}{2.05} & \SIpm{89.92}{3.65} & \SIpm{89.81}{0.86} & \SIpm{89.84}{1.93} & \SIpm{89.21}{2.21} & \SIpm{91.24}{3.32} & \SIpm{90.21}{2.69} \\
\end{longtable}
\end{landscape}

\clearpage
\begin{landscape}
\scriptsize
\setlength{\tabcolsep}{3pt}
\renewcommand{\arraystretch}{1.2}
\begin{longtable}{@{}>{\raggedright\arraybackslash}p{0.17\linewidth}*{6}{>{\centering\arraybackslash}p{0.12\linewidth}}@{}}
\caption{Five-fold cross-validation named entity recognition error breakdown of SciBERT and WaterBERT by entity type.}\label{tab:si8}\\
\toprule
\textbf{Entity Type} & \multicolumn{3}{c}{\textbf{SciBERT}} & \multicolumn{3}{c}{\textbf{WaterBERT}} \\
& \textbf{False Negative Rate (\%)} & \textbf{Partial Match Rate (\%)} & \textbf{Complete Miss Rate (\%)} & \textbf{False Negative Rate (\%)} & \textbf{Partial Match Rate (\%)} & \textbf{Complete Miss Rate (\%)} \\
\midrule
\endfirsthead
\toprule
\textbf{Entity Type} & \multicolumn{3}{c}{\textbf{SciBERT}} & \multicolumn{3}{c}{\textbf{WaterBERT}} \\
& \textbf{False Negative Rate (\%)} & \textbf{Partial Match Rate (\%)} & \textbf{Complete Miss Rate (\%)} & \textbf{False Negative Rate (\%)} & \textbf{Partial Match Rate (\%)} & \textbf{Complete Miss Rate (\%)} \\
\midrule
\endhead
\bottomrule
\endlastfoot
Pollutant & 16.2 & 8.6 & 7.5 & 14.2 & 8.1 & 6.1 \\
Wastewater Treatment Process & 28.3 & 18.6 & 9.7 & 26.1 & 17.1 & 9.0 \\
Reactor & 20.0 & 11.8 & 8.2 & 15.1 & 8.3 & 6.8 \\
Treatment Parameter & 15.4 & 8.2 & 7.2 & 14.1 & 7.4 & 6.8 \\
Microorganism & 18.3 & 11.4 & 6.9 & 16.7 & 9.1 & 7.6 \\
Dosed Material & 27.6 & 15.6 & 12.0 & 23.1 & 14.2 & 8.9 \\
Removal Efficiency & 24.2 & 17.0 & 7.3 & 22.7 & 16.1 & 6.5 \\
Value & 13.3 & 7.9 & 5.5 & 12.6 & 6.8 & 5.7 \\
\end{longtable}
\end{landscape}

\clearpage
\begin{landscape}
\scriptsize
\setlength{\tabcolsep}{4pt}
\renewcommand{\arraystretch}{1.2}
\begin{longtable}{@{}>{\raggedright\arraybackslash}p{0.22\linewidth}*{4}{>{\centering\arraybackslash}p{0.17\linewidth}}@{}}
\caption{Five-fold cross-validation relation extraction error breakdown of SciBERT and WaterBERT by relation type.}\label{tab:si9}\\
\toprule
\textbf{Relation type} & \multicolumn{2}{c}{\textbf{SciBERT}} & \multicolumn{2}{c}{\textbf{WaterBERT}} \\
& \textbf{False Negative Rate (\%)} & \textbf{Mismatch\_Rate (\%)} & \textbf{False Negative Rate (\%)} & \textbf{Mismatch\_Rate (\%)} \\
\midrule
\endfirsthead
\toprule
\textbf{Relation type} & \multicolumn{2}{c}{\textbf{SciBERT}} & \multicolumn{2}{c}{\textbf{WaterBERT}} \\
& \textbf{False Negative Rate (\%)} & \textbf{Mismatch\_Rate (\%)} & \textbf{False Negative Rate (\%)} & \textbf{Mismatch\_Rate (\%)} \\
\midrule
\endhead
\bottomrule
\endlastfoot
\texttt{targets} & 45.4 & 0.4 & 32.8 & 0 \\
\texttt{has\_efficiency} & 38.5 & 0.6 & 35 & 0.3 \\
\texttt{indicates\_removal\_of} & 45.3 & 0.4 & 29.8 & 0.2 \\
\texttt{conditioned\_by} & 34.0 & 0 & 24.7 & 0 \\
\texttt{has\_value} & 11.9 & 0 & 13.3 & 0 \\
\end{longtable}
\end{landscape}

\clearpage
\begin{landscape}
\scriptsize
\setlength{\tabcolsep}{3pt}
\renewcommand{\arraystretch}{1.12}
\begin{longtable}{@{}>{\raggedright\arraybackslash}p{0.16\linewidth}>{\raggedright\arraybackslash}p{0.28\linewidth}>{\raggedright\arraybackslash}p{0.47\linewidth}@{}}
\caption{Examples of graph extraction from different models (highlighted text indicates model extraction errors).}\label{tab:si10}\\
\toprule
\textbf{Source sentence} & \textbf{WaterBERT output} & \textbf{Output of the models that erred} \\
\midrule
\endfirsthead
\toprule
\textbf{Source sentence} & \textbf{WaterBERT output} & \textbf{Output of the models that erred} \\
\midrule
\endhead
\midrule
\multicolumn{3}{r}{Continued on next page}\\
\endfoot
\bottomrule
\endlastfoot
Herein, aiming at efficient adsorption for crystal violet (CV), polyacrylic acid (PAA) was incorporated into zeolitic imidazoline frameworks (ZIF-67) through an easy and environmentally friendly aqueous phase synthesis process at room temperature to synthesize adsorbent ZIF-67-PAA.
& \begin{minipage}[t]{\linewidth}\raggedright
Pollutant: crystal violet (CV); CV; organic dyes\par
\texttt{Wastewater\_Treatment\_Process}: adsorption\par
\texttt{Dosed\_Material}: polyacrylic acid (PAA); zeolitic imidazoline frameworks (ZIF-67); ZIF-67; ZIF-67-PAA
\end{minipage}
& \begin{minipage}[t]{\linewidth}\raggedright
\textbf{Llama-3.3-70B:}\par
Pollutant: crystal violet (CV); CV\par
\texttt{Wastewater\_Treatment\_Process}: adsorption\par
\texttt{Dosed\_Material}: PAA; polyacrylic acid (PAA); zeolitic imidazoline frameworks (ZIF-67); ZIF-67\par
\SIerr{Reactor: ZIF-67-PAA}\par\medskip
\textbf{Qwen3-30B-A3B:}\par
Pollutant: crystal violet; CV; organic dyes\par
\texttt{Wastewater\_Treatment\_Process}: adsorption\par
\texttt{Dosed\_Material}: PAA; polyacrylic acid; zeolitic imidazoline frameworks; ZIF-67\par
Reactor: \SIerr{ZIF-67-PAA}\par
\texttt{Treatment\_Parameter}: adsorption capacity; adsorption performance; Langmuir isotherm model; maximum adsorption capacity; pseudo-second-order kinetic model\par\medskip
\textbf{Qwen3-8B:}\par
Pollutant: crystal violet; dye; organic dyes\par
\texttt{Wastewater\_Treatment\_Process}: adsorption; synthesis\par
\texttt{Dosed\_Material}: aqueous phase; polyacrylic acid; zeolitic imidazoline frameworks\par
Reactor: ZIF-67; \SIerr{ZIF-67-PAA}\par
\texttt{Treatment\_Parameter}: adsorption capacity; aqueous phase synthesis process; maximum adsorption capacity; room temperature
\end{minipage} \\
\addlinespace
Electrostatically precipitated (ESP) recovery boiler fly ash (RBFA), consisting mainly of sodium sulfate Na2SO4, is a solid side stream where HTE are concentrated...The results of this study show that the removal efficiency of the investigated HTE and other analyzed metals was excellent under alkaline conditions, the apparently suitable pH range for the removal of most of these metals being approximately 11.7--12.2.
& \begin{minipage}[t]{\linewidth}\raggedright
Pollutant: Cd; HTE; HTE); Lead; metals; Pb\par
\texttt{Wastewater\_Treatment\_Process}: Electrostatically precipitated (ESP); reverse leaching method\par
\texttt{Dosed\_Material}: sodium sulfate Na2SO4\par
\texttt{Treatment\_Parameter}: liquid solid mass ratio; pH
\end{minipage}
& \begin{minipage}[t]{\linewidth}\raggedright
\textbf{Llama-3.3-70B:}\par
Pollutant: Cd; HTE; Lead; Na2SO4; Pb; \SIerr{sodium sulfate}; Zn\par
\texttt{Wastewater\_Treatment\_Process}: reverse leaching\par
\texttt{Dosed\_Material}: Na2SO4; water\par
\texttt{Treatment\_Parameter}: liquid solid mass ratio; pH\par\medskip
\textbf{Qwen3-30B-A3B:}\par
Pollutant: Cd; hazardous trace elements (HTE); HTE; Lead; Na2SO4; Pb; \SIerr{sodium sulfate}; Zn\par
\texttt{Wastewater\_Treatment\_Process}: reverse leaching method\par
\texttt{Treatment\_Parameter}: liquid solid mass ratio\par\medskip
\textbf{Qwen3-8B:}\par
Pollutant: hazardous trace elements (HTE); Lead; \SIerr{sodium sulfate}\par
\texttt{Wastewater\_Treatment\_Process}: reverse leaching\par
\texttt{Treatment\_Parameter}: degree of removal; energy consumption; liquid solid mass ratio; material losses; pH; removal efficiency; removal rate
\end{minipage} \\
\addlinespace
Comprehensive characterization techniques (XRD, FTIR, SEM, zeta potential, and TGA) confirmed the successful modification of the composites.
& \begin{minipage}[t]{\linewidth}\raggedright
Pollutant: Congo red (CR); methylene blue (MB); organic dyes\par
\texttt{Wastewater\_Treatment\_Process}: Adsorption\par
\texttt{Dosed\_Material}: hexadecyltrimethylammonium bromide (HTAB); montmorillonite; organo-clay nanocomposites; PEG-modified montmorillonite nanocomposites; PEG200; PEG4000; polyethylene glycol (PEG)
\end{minipage}
& \begin{minipage}[t]{\linewidth}\raggedright
\textbf{Llama-3.3-70B:}\par
Pollutant: Congo red; methylene blue; organic dyes\par
\texttt{Wastewater\_Treatment\_Process}: Adsorption\par
\texttt{Dosed\_Material}: hexadecyltrimethylammonium bromide (HTAB); montmorillonite; polyethylene glycol (PEG)\par
Reactor: nanocomposites\par
\texttt{Treatment\_Parameter}: basal spacing; kinetics; molecular weights; \SIerr{zeta potential}\par\medskip
\textbf{Qwen3-30B-A3B:}\par
Pollutant: Congo red; methylene blue; organic dyes\par
\texttt{Wastewater\_Treatment\_Process}: Adsorption\par
\texttt{Dosed\_Material}: hexadecyltrimethylammonium bromide (HTAB); PEG200; PEG200/MMT@HTAB; PEG4000; PEG4000/MMT@HTAB\par
\texttt{Treatment\_Parameter}: basal spacing d; \SIerr{zeta potential}\par\medskip
\textbf{Qwen3-8B:}\par
Pollutant: Congo red (CR); methylene blue (MB); organic dyes\par
\texttt{Wastewater\_Treatment\_Process}: Adsorption\par
\texttt{Dosed\_Material}: hexadecyltrimethylammonium bromide (HTAB); HTAB; MMT@HTAB; montmorillonite; PEG200; PEG4000\par
Reactor: MMT@HTAB\par
\texttt{Treatment\_Parameter}: adsorption mechanism; basal spacing; characterization techniques; dipole interactions; FTIR; interlayer expansion; molecular weight; physical adsorption; pore network; porous framework; SEM; TGA; wastewater treatment; XRD; \SIerr{zeta potential}
\end{minipage} \\
\addlinespace
The MoO2/C spheres act as cathodes of an asymmetric CDI device and effectively reduce the concentration of Pb2+ from 50 ppm to $<0.21$ ppb.
& \begin{minipage}[t]{\linewidth}\raggedright
Pollutant: heavy metal pollutant; lead; lead ions; Pb(NO3)(2); Pb2+\par
\texttt{Wastewater\_Treatment\_Process}: asymmetric capacitive deionization (CDI); asymmetric CDI; SCR; selective capacitive removal (SCR)\par
\texttt{Dosed\_Material}: MoO2/C; NaCl
\end{minipage}
& \begin{minipage}[t]{\linewidth}\raggedright
\textbf{Llama-3.3-70B:}\par
Pollutant: heavy metal; heavy metal ions; lead; lead ions; Pb2+\par
\texttt{Wastewater\_Treatment\_Process}: capacitive deionization; selective capacitive removal\par
\texttt{Dosed\_Material}: [MoO4]; [MoO6]; molybdenum dioxide/carbon; MoO2/C; MoO2/C electrodes; NaCl; Pb(NO3)(2)\par
Reactor: asymmetric CDI device\par
\texttt{Treatment\_Parameter}: concentration; intercalation; \SIerr{ppb; ppm}; regeneration performance; SCR efficiency\par\medskip
\textbf{Qwen3-30B-A3B:}\par
Pollutant: heavy metal; lead\par
\texttt{Wastewater\_Treatment\_Process}: capacitive deionization; selective capacitive removal\par
\texttt{Dosed\_Material}: MoO2/C\par
Reactor: MoO2/C electrodes\par
Microorganism: [MoO4]; [MoO6]\par
\texttt{Treatment\_Parameter}: NaCl; Pb(NO3)(2); Pb2+; ppb; \SIerr{ppm}\par\medskip
\textbf{Qwen3-8B:}\par
Pollutant: heavy metal ions; lead; lead ions; NaCl; Pb(NO3)(2); Pb2+\par
\texttt{Wastewater\_Treatment\_Process}: capacitive deionization; selective capacitive removal\par
\texttt{Dosed\_Material}: MoO2/C\par
Reactor: CDI\par
\texttt{Treatment\_Parameter}: concentration; efficiency; \SIerr{ppb; ppm;} regeneration performance
\end{minipage} \\
\end{longtable}
\end{landscape}

\setcounter{figure}{\SIoldfigure}
\setcounter{table}{\SIoldtable}
\setcounter{equation}{\SIoldequation}
\endgroup